\documentclass[sn-mathphys-num]{sn-jnl}

\usepackage{graphicx}%
\usepackage{multirow}%
\usepackage{amsmath,amssymb,amsfonts}%
\usepackage{amsthm}%
\usepackage{mathrsfs}%
\usepackage[title]{appendix}%
\usepackage{xcolor}%
\usepackage{textcomp}%
\usepackage{manyfoot}%
\usepackage{booktabs}%
\usepackage{algorithm}%
\usepackage{algorithmicx}%
\usepackage{algpseudocode}%
\usepackage{listings}%
\usepackage{rotating}
\usepackage{makecell}

\usepackage{xr-hyper}
\makeatletter
\IfFileExists{2_supplementary.aux}{%
  \externaldocument[supp-]{2_supplementary}%
}{}
\makeatother

\usepackage{amsmath}
\usepackage{amssymb}
\usepackage{accents}
\usepackage{mathtools}
\usepackage{makecell}
\usepackage{multirow}
\usepackage{caption}
\usepackage{fullpage}

\usepackage{longtable}
\usepackage{booktabs}
\usepackage{lscape}
\usepackage{makecell}
\usepackage{array}
\usepackage{geometry}

\usepackage{booktabs}
\usepackage{longtable}
\usepackage{makecell}

\usepackage{color,soul}
\usepackage{xcolor}
\definecolor{cerulean}{rgb}{0.0,0.48,0.65}
\definecolor{green}{rgb}{0.01, 0.75, 0.24}
\definecolor{Black}{RGB}{0.0, 0.0, 0.0}
\newcommand{\red}[1]{\textcolor{Black}{#1}}
\newcommand{\blue}[1]{\textcolor{blue}{#1}}

\newcommand{\brown}[1]{\textcolor{Black}{#1}}

\newcommand{\teal}[1]{\textcolor{teal}{#1}}

\newcommand{\shadow}[1]{}
\renewcommand{\arraystretch}{1.15}
\def\r{\red}
\def\b{\blue}

\def\s{\shadow}

\def\br{\brown}

\usepackage{pifont}
\newcommand{\cmark}{\ding{51}}%
\theoremstyle{thmstyleone}%
\theoremstyle{thmstyletwo}%

\theoremstyle{thmstylethree}%

\begin{document}
\pagestyle{empty}
\thispagestyle{empty}

\title[Article Title]{Foundation Models for Epileptogenic Zone Identification in Drug-Resistant Epilepsy}


\author[1,2]{\fnm{Thi Kieu Khanh} \sur{Ho}}\s{\email{thi.k.ho@mail.mcgill.ca}}
\equalcont{These authors contributed equally to this work.}

\author[1,2]{\fnm{Thomas} \sur{Lai}}\s{\email{thomas.lai@mail.mcgill.ca}}
\equalcont{These authors contributed equally to this work.}

\author[3,4,7]{\fnm{Petr} \sur{Klimes}}

\author[5,6]{\fnm{Jan} \sur{Cimbalnik}}

\author[5,6]{\fnm{Martin} \sur{Pail}}

\author[5,6]{\fnm{Milan} \sur{Brazdil}}


\author[3,7,8]{\fnm{Birgit} \sur{Frauscher}}

\author*[1,2]{\fnm{Narges} \sur{Armanfard}}\email{narges.armanfard@mcgill.ca}

\affil[1]{\orgdiv{Department of Electrical and Computer Engineering}, \orgname{McGill University}, \city{Montréal, Quebec}, \country{Canada}}

\affil[2]{\orgname{Mila-Quebec AI Institute}, \city{Montréal, Quebec}, \country{Canada}}

\affil[3]{\orgname{Montreal Neurological Institute and Hospital, McGill University}, \city{Montréal, Quebec}, \country{Canada}}

\affil[4]{\orgname{Institute of Scientific Instruments, The Czech Academy of Sciences}, \city{Brno}, \country{Czech Republic}}

\affil[5]{\orgname{Brno Epilepsy Center, Department of Neurology, St Anne's University Hospital}, \city{Brno}, \country{Czech Republic}}

\affil[6]{\orgname{Faculty of Medicine, Masaryk University}, \city{Brno}, \country{Czech Republic}}



\affil[7]{\orgname{Analytical Neurophysiology Lab, Department of Neurology, Duke University Medical Center, Durham, NC}, \country{USA}}

\affil[8]{\orgname{Department of Biomedical Engineering, Duke Pratt School of Engineering, Durham, NC}, \country{USA}}

\abstract{Accurate identification of the epileptogenic zone (EZ) is essential for seizure freedom after resective surgery in drug-resistant epilepsy, yet \br{\s{complete }seizure freedom}\s{current surgical success} rates remain below 50\%. We developed EpiiSLM, a dual foundation model system for EZ identification with stereo-electroencephalography (sEEG), by training a signal foundation model on 104{,}990 minutes of sEEG recordings from the Montreal Neurological Institute \& Hospital, \br{while leveraging all recordings regardless of surgical outcome and anchoring EZ biomarker extraction on non-epileptic signals.}\s{using all available patients during unsupervised learning and anchoring supervised fine-tuning on a nonambiguous non-epileptic class.} A language foundation model then integrates sEEG-derived outputs with multimodal clinical information to produce interpretable predictions. Under leave-one-patient-out evaluation, EpiiSLM achieved 0.978 contact-level \br{positive predictive value (PPV)}\s{PPV$_3$}, outperforming \br{the \s{naïve }seizure onset zone(SOZ)-as-EZ} baseline by 15.1\% ($p<0.05$), and 100\% region-level accuracy; on an external dataset, EpiiSLM achieved 0.857 contact-level \br{PPV}\s{PPV$_3$}. EpiiSLM requires only one night of interictal sleep data, suggesting potential to reduce invasive sEEG monitoring duration and improve surgical outcomes.}




\maketitle
 
\section{Introduction}\label{introduction}

Epilepsy affects tens of millions of individuals worldwide, who face sudden, unpredictable seizures that carry a high risk of injury, suffocation, and death \cite{surges2021identifying}. Although seizures can be controlled with medication in most patients, approximately one-third are drug-resistant \cite{jamiolkowski2024fasciola,rahim2021experimental}. For these patients, the standard treatment is surgical resection of the epileptogenic zone (EZ)--the brain region responsible for seizure generation \cite{wiebe2012pharmacoresistance}\s{engel2012early,}. \br{Accurate identification and complete resection of the EZ can render patients seizure-free (Engel Ia \cite{engel1993update}/ILAE 1 \cite{wieser2001ilae}), yet 50--60\% of patients who undergo EZ resection continue to experience seizures despite irreversible \r{removal}\s{loss} of brain tissue; and many require repeated surgery, each carrying additional surgical risk\s{, before achieving seizure freedom}  \cite{Avigdorjnnp-2025-337158,malmgren2017long,jaber2024spatial}, underscoring the need for more accurate EZ localization.}\s{Successful identification and resection of the EZ can result in complete seizure freedom, yet 50-60\% of drug-resistant patients who undergo EZ resection continue to experience seizures despite losing brain tissue \cite{Avigdorjnnp-2025-337158,malmgren2017long,jaber2024spatial}, motivating the need for more accurate EZ identification methods.}

Most existing computational methods, including artificial intelligence (AI) methods, rely on magnetic resonance imaging (MRI) \cite{klimes2022spatio}\s{grouiller2011or, }, positron emission tomography (PET) \cite{casse2002positron}, scalp electroencephalography (EEG) \cite{ramantani2016correlation}, or \br{\r{intracranial EEG (iEEG)}\s{electrocorticography (ECoG)} \cite{ds005398:1.1.1,ds007118:1.0.0},} incorporating indicators such as \s{Do we remove pathology? Birgit didn't point it out, but Jean did}pathology \cite{jehi2018epileptogenic}, brain lesions \cite{luders2006epileptogenic}, and the seizure onset zone (SOZ), which are the regions where seizures are observed to begin \cite{luders2006epileptogenic}. However, these modalities lack the temporal and spatial resolution needed for precise EZ localization. Neurologists consider that the most informative biomarkers would be found with \br{stereo-electroencephalography (sEEG) \cite{mullin2016seeg}, a type of \r{iEEG}\s{intracranial EEG (iEEG)} characterized by deep intracerebral electrodes, }\s{intracranial EEG, particularly stereo-electroencephalography (sEEG), }despite the fact that sEEG is invasive, requires hospitalization, and demands a specialized team of neurosurgeons and neurophysiologists for both acquisition and interpretation.

Consequently, AI models for sEEG-based EZ identification \cite{klimes2019nrem,balaji2024seizure,nejedly2025leveraging,klimes2024interictal} are scarce and face two key challenges. \br{First, previous EZ-identification methods underutilize most sEEG data, because they rely on manually selected intervals (10--30 minutes per patient) and include only patients with good surgical outcomes \cite{nejedly2025leveraging,klimes2019nrem,klimes2022spatio}.}\s{First, because surgical success rates remain low, only a small fraction of available recordings come from seizure-free-post-surgery patients \cite{klimes2019nrem,balaji2024seizure}, leaving most sEEG data underutilized in conventional supervised learning.} \br{Second, because precise EZ labels are currently unobtainable even after surgery, prior work approximated EZ contacts (conductive points between sEEG electrodes and brain tissue) as resected \cite{makaram2023deep,nejedly2025leveraging} or seizure onset zone (SOZ) \cite{abdallah2026quantifying} contacts, or their intersection \cite{klimes2019nrem}.}\s{Second, no reliable method exists to \br{precisely label EZ contacts}\s{retrospectively localize the EZ} even when surgical outcome is known, making \br{label-dependent model training difficult}\teal{(for now)}\s{ground truth EZ labels effectively unobtainable} . In sEEG-based EZ identification, contacts, or conductive points between sEEG electrodes and brain tissue, serve as the spatial unit of analysis. Prior work has approximated EZ contacts using resected contacts \cite{makaram2023deep,nejedly2025leveraging}, SOZ contacts \cite{balaji2022seizure}, or their intersection \cite{klimes2019nrem}.} 
These approximations are imprecise, as resection boundaries are subjective and often include non-EZ tissue, and resecting the SOZ alone does not guarantee seizure freedom \cite{luders2006epileptogenic,makaram2023deep,abdallah2026quantifying}. Models trained or evaluated against such labels may learn biomarkers unrelated to the EZ, leading to high bias and false positives, which are clinically associated with functional loss due to \br{model-misguided}\s{the} resection of healthy tissues.

To address these challenges, we introduce EpiiSLM (\textbf{Epi}leptogenic zone \textbf{i}dentification \textbf{S}ignal-\textbf{L}anguage \textbf{M}odel), which to our knowledge is the first dual-foundation model system for EZ identification. A signal foundation model (SFM) learns transferable representations from the largely untapped sEEG data source through unsupervised learning. Because it extracts biomarkers exclusively from interictal (non-seizure) recordings, EpiiSLM reduces the need to prolong sEEG monitoring to capture ictal events, potentially shortening hospitalization from up to two weeks \cite{klimes2019nrem} to a single overnight session. To resolve ground truth ambiguity, we anchor the learning objective on a class that can be defined without doubt: the negative \emph{non-epileptic} class defined as non-resected, non-SOZ contacts in seizure-free-post-surgery patients\s{--contacts that}, which, by definition, cannot contain any EZ. Drawing from one-class classification \cite{tang2021self,ho2023self,ho2023multivariate,lai2023open}, EpiiSLM models the distribution of this negative class and identifies candidate EZ contacts as deviations from it, shifting the anchor from ambiguous positive labels to a grounded negative class. Finally, a language foundation model (LFM) incorporates \emph{a priori} information from other modalities such as SOZ locations and MRI findings into the final EZ predictions while providing natural-language reasoning to maximize interpretability. Together, these contributions move EZ identification toward higher accuracy, lower bias, and greater explainability, with the potential to improve resective-surgery outcomes in drug-resistant epilepsy.

\section{Results}\label{results}

We trained EpiiSLM on 104,990 minutes (868 GB) of sEEG data from 30 drug-resistant epilepsy patients (17 female / 13 male) who underwent resective surgery at the Montreal Neurological Institute \& Hospital (MNI) in Montreal, Quebec, Canada, between 2010 and 2017. Given the complexity of invasive sEEG data collection, our strict inclusion criteria, and low post-surgical seizure-freedom rates, this dataset, which is comparable in size to those used in prior studies \cite{klimes2019nrem, makaram2023deep, balaji2024seizure, jamiolkowski2024fasciola}, is among the most diverse available for this task in terms of patient age, underlying pathology, electrode type, number of contacts, and spatial coverage (Table \ref{tab:clinical-characteristics}). We further evaluated EpiiSLM's generalizability on an external cohort of 17 patients from St Anne's University Hospital (Brno) in Brno, Czech Republic, for a total of 47 patients. \br{Although public iEEG datasets exist \cite{ds005398:1.1.1,ds007118:1.0.0}, to our knowledge, no sEEG dataset of comparable size, that fits our experimental protocol and provides the scale required for foundation-model training, is currently available, owing to the difficulty of \r{curating long term}\s{acquiring} invasive recordings and privacy constraints on sharing clinical data \cite{rahimzadeh2023benefits}.}\s{No comparable large-scale public sEEG dataset currently exists for epileptogenic zone (EZ) identification, owing to the difficulty of \r{curating long-term}\s{acquiring} invasive recordings and privacy constraints on sharing clinical data \cite{rahimzadeh2023benefits}.}

 EpiiSLM comprises two main components: a trainable SFM--to our knowledge, the first sEEG foundation model developed for EZ identification--and a frozen LFM that integrates clinical information and produces medical reasoning. We trained the SFM in two phases analogous to the pre-training and fine-tuning stages of large language models \cite{kenton2019bert}.  In Training Phase I, we used an unsupervised similarity learning objective with masking \cite{biot} to learn a lower-dimensional representation of sEEG signals. Unlike existing methods  \cite{klimes2019nrem,makaram2023deep} that discarded patients with poor surgical outcomes (Engel classes IIa-IVb \cite{engel1993update}), we incorporated their data by training on an auxiliary task that requires no ground-truth labels, thereby improving downstream generalization. In Training Phase II, we used resection, SOZ, and surgical outcome ground truths to train two complementary EZ-identification heads: a One-Class Head and a Binary Head, both grounded on the well-defined non-epileptic class.  We combined these heads as an ensemble to produce preliminary, contact-level EZ scores. The LFM then integrates SOZ, MRI, and clinical priors with these scores to produce final predicted EZ contacts and regions with medical reasoning, by leveraging language models pre-trained on medical question-answering \cite{sellergren2025medgemma,singhal2025toward} and prompt engineering techniques \cite{liu2023large,wei2022chain}. An overview of EpiiSLM's framework is shown in Fig.~\ref{fig:model}.

\subsection{EpiiSLM achieves high-precision epileptogenic zone identification}\label{sec:general_performance} 

Because direct EZ ground truth is currently unobtainable, we used the resected contacts in patients who were seizure-free after surgery as a proxy for positives during model evaluation. Under this approximation, false negatives are ill-defined: a resected contact that receives a low predicted EZ probability does not establish whether removing that contact was necessary for seizure freedom. False positives, by contrast, carry direct clinical risk, as they may encourage unnecessary resection and subsequent functional deficits. We therefore adopted positive predictive value (PPV; also termed precision) as the primary contact-level metric, which penalizes false positives without relying on the problematic false-negative assumption in area under the receiver-operating curve (AUROC) and area under the precision-recall curve (AUPRC)--metrics that have been prioritized for this task \cite{hrtonova2025metrics,nejedly2025leveraging,klimes2019nrem} despite the fact that optimizing them with approximate positives can yield biased models (Supplementary Note \b{1}). We define two PPV variants: PPV$_0$ measures strict precision against the resection boundary, while PPV$_3$ extends that boundary by three contacts ($\sim$10.5 mm) to accommodate the spatial sensitivity of sEEG electrodes and potential co-registration error \cite{VONELLENRIEDER20211105,nejedly2025leveraging,dessert2023optimization,reinacher2025combined}.
 
We evaluated EpiiSLM under a leave-one-patient-out protocol across all completely seizure-free patients (Engel Class Ia \cite{engel1993update}) (Table~\ref{tab:compare_results}). EpiiSLM achieved a mean PPV$_0$ of 0.777 (95\% CI 0.489--0.936) (specificity $=0.982\pm0.026$; AUROC $=0.939\pm0.064$; AUPRC $=0.707\pm0.242$), 5.9\% higher than the naïve SOZ-as-EZ baseline, in which the predicted EZ consists directly of the SOZ contacts. EpiiSLM achieved a mean PPV$_3$ of 0.978 (95\% CI 0.897--1.000), a 15.1\% improvement over the same baseline ($p<0.05$, Wilcoxon signed-rank test). AUROC and AUPRC are reported for completeness but should be interpreted with caution given the ambiguity of false negatives in this setting. Per-run results are provided in Supplementary Note \b{6}. Although direct numerical comparison to prior work is precluded by differences in cohorts and evaluation protocols, reported PPV$_0$ values in the literature range from approximately 30\% \cite{makaram2023deep} to 64\% \cite{nejedly2025leveraging} with respect to resected contacts, and 58\% with respect to the intersection of resected and SOZ contacts \cite{klimes2019nrem}. Qualitative examples, including region-level EZ predictions, are discussed in Section~\ref{sec:qualitative}.

\newgeometry{margin=1.5cm}
\setlength{\tabcolsep}{0.8pt}
\renewcommand{\arraystretch}{2.0}
\setlength{\LTleft}{0pt plus 1fill}
\setlength{\LTright}{0pt plus 1fill}

\begin{landscape}
\footnotesize
\begin{longtable}{ccccccccccccc}

\caption{\small{\textbf{Clinical characteristics of the Montreal Neurological Institute and Hospital study cohort.}}
\textbf{Epilepsy type:} FIE, fronto-insular epilepsy; FLE, frontal lobe epilepsy; OIE, operculo-insular epilepsy; PLE, parietal lobe epilepsy; PQE, posterior quadrant epilepsy; TIE, temporo-insular epilepsy; TLE, temporal lobe epilepsy; TOE, temporo-occipital epilepsy.
\textbf{Pathology, MRI lesion, interictal sEEG and resected regions:} FCD, focal cortical dysplasia; L, left; PNH, periventricular nodular heterotopia; R, right; WHO, World Health Organization.
\textbf{SOZ contacts and resected contacts:} A, amygdala; AG, angular gyrus; CA, anterior cingulate; CM, mid-cingulate; CP, posterior cingulate; FP, frontopolar; FS, superior frontal; FUG, fusiform gyrus; H, hippocampus; HA, anterior hippocampus; HE, Heschl's gyrus; HP, posterior hippocampus; IA, anterior insula; IP, posterior insula; L, left; LA, anterior lesion; LE, lesion; LG, lingual gyrus; LI, inferior lesion; LP, posterior lesion; LS, superior lesion; OF, orbitofrontal; OI, inferior occipital; OS, superior occipital; R, right; TA, anterior temporal; TM, mid-temporal; TP, temporal pole; TR, trigonum.}\label{tab:clinical-characteristics}

\\

\toprule
\makecell{\textbf{Patient} \\ \textbf{ID}} & \textbf{Age} & \textbf{Gender} & \makecell{\textbf{Epilepsy} \\ \textbf{type}} & \textbf{Pathology} & \textbf{MRI lesion} & \makecell{\textbf{Interictal} \\ \textbf{sEEG}} & \makecell{\textbf{Contacts} \\ \textbf{implanted}} & \makecell{\textbf{SOZ} \\ \textbf{contacts} (\textit{n})} & \makecell{\textbf{Resective} \\ \textbf{surgery}} & \makecell{\textbf{Resected} \\ \textbf{contacts} (\textit{n})} & \makecell{\textbf{Follow-up} \\ \textbf{(months)}} & \textbf{Outcome} \\
\midrule
\endfirsthead

\multicolumn{13}{l}{\textbf{Table~\ref{tab:clinical-characteristics} (continued)}} \\[4pt]
\toprule
\makecell{\textbf{Patient} \\ \textbf{ID}} & \textbf{Age} & \textbf{Gender} & \makecell{\textbf{Epilepsy} \\ \textbf{type}} & \textbf{Pathology} & \textbf{MRI lesion} & \makecell{\textbf{Interictal} \\ \textbf{sEEG}} & \makecell{\textbf{Contacts} \\ \textbf{implanted}} & \makecell{\textbf{SOZ} \\ \textbf{contacts} (\textit{n})} & \makecell{\textbf{Resective} \\ \textbf{surgery}} & \makecell{\textbf{Resected} \\ \textbf{contacts} (\textit{n})} & \makecell{\textbf{Follow-up} \\ \textbf{(months)}} & \textbf{Outcome} \\
\midrule
\endhead

\bottomrule
\endlastfoot

1 & 26 & Female & FLE & FCD 2A & Normal & \makecell{R orbitofrontal lateral to \\ mid-frontal convexity and \\ anterior temporal neocortex} & 90 & ROF6--8 (3) & \makecell{R orbitofrontal \\ lateral resection} & ROF1--12 (12) & 29 & Ia \\ \midrule

2 & 28 & Female & FLE & FCD 2B & \makecell{L second frontal gyrus \\ FCD} & L second frontal gyrus (lesion) & 55 & \makecell{LCM6--8, \\ LFS4--8 (8)} & \makecell{L second frontal gyrus \\ lesionectomy} & \makecell{LCM5--6, \\ LFS3--5 (5)} & 12 & Ia \\ \midrule

3 & 22 & Male & TLE & \makecell{Hippocampal \\ gliosis} & \makecell{R hippocampal sclerosis; \\ L occipital ulegyria/ \\ encephalomalacia} & \makecell{R mesiotemporal \\ and neocortex} & 83 & \makecell{RA1--3, RH1--4, \\ RHP1--3 (10)} & \makecell{R selective amygdalo-\\hippocampectomy} & \makecell{RA1--10, \\ RH1--3 (13)} & 65 & Ia \\ \midrule

4 & 38 & Male & FLE & FCD 2B & Normal & R orbitofrontal & 95 & ROF1--15 (15) & \makecell{R orbitofrontal \\ resection} & ROF5--15 (11) & 59 & Ia \\ \midrule
 
5 & 33 & Male & FLE & FCD 2A & \makecell{R mid-frontal convexity \\ FCD} & \makecell{R mid-frontal convexity \\ (lesion)} & 110 & RLI1--7 (7) & \makecell{R mid-frontal convexity \\ lesionectomy} & \makecell{RLI1--7, RLS1--3, \\ RLS6--7 (12)} & 54 & Ia \\ \midrule

6 & 53 & Female & TLE & FCD 2A & Normal & L mesial temporal & 91 & \makecell{LA1--3, LH1--3, \\ LHP1--3, \\ LTM1--3 (12)} & \makecell{L selective amygdalo-\\hippocampectomy} & \makecell{LA1--8, LH1--3, \\ LTA1--7 (18)} & 47 & Ia \\ \midrule

7 & 21 & Female & PLE & FCD 2B & R pre-cuneus FCD & R pre-cuneus (lesion) & 90 & \makecell{RLA1--10, \\ RLP1--8 (18)} & \makecell{R pre-cuneus \\ lesionectomy} & \makecell{RLA1--10, RLP1--8, \\ RAG1--4 (22)} & 41 & Ia \\ \midrule

8 & 36 & Male & \makecell{Bilateral \\ TLE} & PNH & Bilateral focal PNH & \makecell{Bilateral temporal widespread \\ L to R and \\ heterotopic tissue} & 115 & \makecell{LA6--9, LFUG4--9, \\ LHA4--9, LHP4--9, \\ RHA1--6, RHP1--6 (34)} & \makecell{L anterior temporal \\ lobectomy} & \makecell{LA3--13, LHA1--11, \\ LHP1--10 (32)} & 25 & Ia \\ \midrule

9 & 27 & Female & TLE & FCD 2A & Normal & \makecell{R temporal neocortex $>$ \\ orbitofrontal region \\ and anterior insula} & 85 & \makecell{RTA3--5, RA6--8, \\ RH6--8 (9)} & \makecell{R anterior temporal \\ lobectomy} & \makecell{RTA1--5, RA1--8, \\ RH2--8 (20)} & 36 & Ia \\ \midrule

10 & 36 & Male & FLE & FCD 2B & \makecell{L anterior cingulate \\ FCD} & \makecell{L anterior cingulate and \\ orbitofrontal; L hippocampus} & 102 & \makecell{LCA1--3, \\ LOF1--3 (6)} & \makecell{L anterior cingulate \\ lesionectomy} & LCA1--3 (3) & 42 & Ia \\ \midrule

11 & 32 & Female & TOE & FCD 3D & \makecell{L posterior cerebral arterial \\ territory remote ischemic lesion; \\ R posterior cerebral arterial \\ territory discrete lesion} & \makecell{L inferior \\ parietal-temporal-occipital} & 88 & \makecell{LFUG1--3, \\ LHP1--4 (7)} & \makecell{L anterior temporal \\ lobectomy} & \makecell{LHP1--3, LA1--8, \\ LTP1--5 (16)} & 60 & Ib \\ \midrule

12 & 14 & Male & FLE & FCD 2B & \makecell{R pre-cuneus, mesial \\ parietal and parietal \\ cingulate cystic lesion; \\ R orbitofrontal and \\ anterior cingulate FCD} & \makecell{R orbitofrontal and frontal \\ convexity $>$ supplementary \\ motor area and cingulate \\ gyrus; parietal in the vicinity \\ of the surgical cavity} & 114 & ROF1--6 (6) & \makecell{R orbitofrontal \\ resection \\ (lesionectomy)} & \makecell{ROF1--15, \\ RFP1--11 (26)} & 30 & Ib \\ \midrule

13 & 56 & Male & \makecell{Bilateral \\ TLE} & FCD 2A & \makecell{R frontal encephalomalacia; \\ bilateral hippocampal atrophy} & \makecell{R mesiotemporal and \\ orbitofrontal $>>$ L mesial and \\ neocortical temporal} & 95 & \makecell{LA1--3, LH1--3, \\ RA1--3, RH1--3 (12)} & \makecell{R selective amygdalo-\\hippocampectomy} & \makecell{RA1--3, \\ RH1--3 (6)} & 37 & Ib \\ \midrule

14 & 38 & Female & TLE & FCD 2A & \makecell{L orbitofrontal encephalocele; \\ L hippocampal malrotation} & \makecell{L mesial and \\ neocortical temporal} & 59 & \makecell{LA1--3, LH1--7, \\ LHP1--4, LOF4--5 (16)} & \makecell{L anterior temporal \\ lobectomy sparing \\ the hippocampus} & \makecell{LA4--12, \\ LH1 (10)} & 64 \\ \hline

\pagebreak


15 & 39 & Male & \makecell{Bilateral \\ TLE} & \makecell{Hippocampal \\ sclerosis} & R hippocampal atrophy & \makecell{Bilateral temporal neocortical \\ and mesial; bilateral orbitofrontal} & 114 & \makecell{LH1--8, LHP1--9, \\ LA1--9, RA1--9, \\ RH1--9, RHP1--9 (53)} & \makecell{R selective amygdalo-\\hippocampectomy} & \makecell{RH1--2, RA1--3, \\ RA8--9 (7)} & 43 & Id \\ \midrule

16 & 38 & Male & TOE & FCD 1B & \makecell{L posterior insula, temporal \\ and posterior temporal, \\ and inferior parietal atrophy \\ and gliosis} & \makecell{Multifocal L lingual gyrus, \\ cuneus, posterior hippocampus, \\ and pre-cuneus} & 95 & \makecell{LHP1--3, LL3--8, \\ LOS1--3 (12)} & \makecell{Partial L occipital \\ lobe resection} & LOS1--4 (4) & 34 & IIa \\ \midrule

17 & 25 & Female & TLE & \makecell{Hippocampal \\ sclerosis} & \makecell{L mesiotemporal sclerosis \\ and surgical bed (L temporal \\ pole and amygdala)} & \makecell{Residual L hippocampus \\ to posterior cingulate gyrus} & 67 & \makecell{LCP1--2, LH2--3, \\ LHP1--2 (6)} & \makecell{Residual L \\ hippocampal resection} & LH1--4 (4) & 86 & IIIa \\ \midrule

18 & 33 & Female & TOE & Gliosis & Normal & \makecell{Widespread L temporo-occipital \\ and R mesiotemporal} & 89 & \makecell{LA1--3, LH1--3, \\ LFUG1--4, \\ LOI8--14 (17)} & \makecell{L temporo-occipital \\ resection} & LOI11--14 (4) & 27 & IIIa \\ \midrule

19 & 30 & Female & TLE & \makecell{Hippocampal \\ gliosis} & L hippocampal atrophy & \makecell{L mesiotemporal and \\ fusiform $>$ temporal neocortex} & 88 & \makecell{LFUG1--3, LH1--3, \\ LHP1--3, LA1--3 (12)} & \makecell{L anterior temporal \\ lobectomy} & \makecell{LH1--8, \\ LA1--8 (16)} & 15 & IIIa \\ \midrule

20 & 34 & Male & \makecell{Bilateral \\ TLE} & FCD 2A & Normal & \makecell{Bilateral mesial and neocortical \\ temporal, L$>>$R} & 74 & \makecell{LA1--3, LH1--8, \\ LHP1--6, RH1--3, \\ RHP1--3 (23)} & \makecell{L anterior temporal \\ lobectomy} & LA1--8 (8) & 13 & IIIa \\ \midrule

21 & 25 & Female & \makecell{Bilateral \\ TLE} & \makecell{Hippocampal \\ gliosis} & Normal & Bilateral mesiotemporal & 77 & LH1--4 (4) & \makecell{L anterior temporal \\ lobectomy} & \makecell{LH1--8, \\ LA1--8 (16)} & 12 & IIIb \\ \midrule

22 & 30 & Male & OIE & Gliosis & \makecell{L centro-parietal \\ and posterior insular \\ encephalomalacia} & \makecell{L posterior insula \\ $>$ parietal} & 79 & \makecell{LCP1--8, \\ LIP1--4 (12)} & \makecell{L inferior parietal-\\ insular resection} & LIP1--5 (5) & 43 & IIIb \\ \midrule

23 & 42 & Female & TLE & \makecell{Hippocampal \\ gliosis} & \makecell{L hippocampal atrophy \\ and signal change} & Mesiotemporal, L$>$R & 88 & \makecell{LA1--3, LH1--3, \\ LHP2--3 (8)} & \makecell{L selective amygdalo-\\hippocampectomy} & \makecell{LA1--12, \\ LH1--11 (23)} & 20 & IVa \\ \midrule

24 & 30 & Male & FIE & FCD 2B & Normal & \makecell{Widespread L frontal \\ with anterior insula} & 116 & LI1--7, LOF1--7 (14) & \makecell{L orbitofrontal \\ resection} & LOF1--11 (11) & 34 & IVa \\ \midrule

25 & 52 & Female & \makecell{TLE \\ plus} & \makecell{Ganglioglioma \\ WHO grade I} & \makecell{L Heschl gyrus \\ and posterior \\ insula FCD} & \makecell{L hippocampus and \\ anterior and mid-temporal \\ neocortex} & 55 & \makecell{LHE3--6, \\ LIP3--6 (8)} & \makecell{L Heschl gyrus \\ resection} & LHE1--6 (6) & 46 & IVa \\ \midrule

26 & 29 & Female & FLE & FCD 2B & \makecell{R orbitofrontal \\ hypersignal} & \makecell{R orbitofrontal, frontopolar \\ and mesiotemporal} & 100 & ROF1--3 (3) & \makecell{R orbitofrontal \\ cortectomy} & \makecell{RL1--10, RFP1--10, \\ ROF3--10 (28)} & 67 & IVa \\ \midrule

27 & 21 & Female & PQE & \makecell{PNH; FCD \\ not further \\ specified} & \makecell{Bilateral temporo-parieto-occipital \\ PNH, abnormal gyration L$>$R \\ inferior posterior temporal region \\ and cerebellar dysgenesis; \\ R temporal encephalomalacia \\ (post-trauma)} & \makecell{Widespread temporo-parietal \\ mesial and neocortical, R$>$L, \\ and heterotopic tissue} & 118 & \makecell{RTR3--8, RLP3--7, \\ RLE3--8, \\ RHP3--8 (23)} & \makecell{R posterior inferior \\ temporal resection} & \makecell{RLP1--7, RLE3--9, \\ RHP2--8, \\ RH3--5 (24)} & 35 & IVa \\ \midrule

28 & 38 & Female & \makecell{TLE} & Gliosis & Normal & Mesiotemporal, L$>>>$R & 66 & \makecell{LA1--3, LH1--3, \\ LHP1--3 (9)} & \makecell{L selective amygdalo-\\hippocampectomy} & \makecell{LA1--9, \\ LH1--4 (13)} & 41 & IVa \\ \midrule

29 & 27 & Female & TIE & \makecell{Probably \\ FCD} & \makecell{L Heschl gyrus and \\ L posterior insula FCD} & \makecell{L mid-posterior temporal \\ neocortex $>$ Heschl gyrus, \\ transverse gyrus and \\ posterior insula (lesion)} & 117 & \makecell{LIP4--6, \\ LHE1--5 (8)} & \makecell{L Heschl gyrus \\ resection} & LHE3--6 (4) & 35 & IVb \\ \midrule

30 & 24 & Female & TLE & Gliosis & \makecell{R hippocampal hypersignal; \\ L frontal cystic-like lesion} & R mesiotemporal & 84 & \makecell{RA1--3, RH1--6, \\ RHP1--10 (19)} & \makecell{R selective amygdalo-\\hippocampectomy} & \makecell{RA1--11, \\ RH1--13 (24)} & 51 & IVb \\ 

\end{longtable}
\end{landscape}
\restoregeometry

\begin{figure}[H]
    \centering
    \includegraphics[width=1\linewidth]{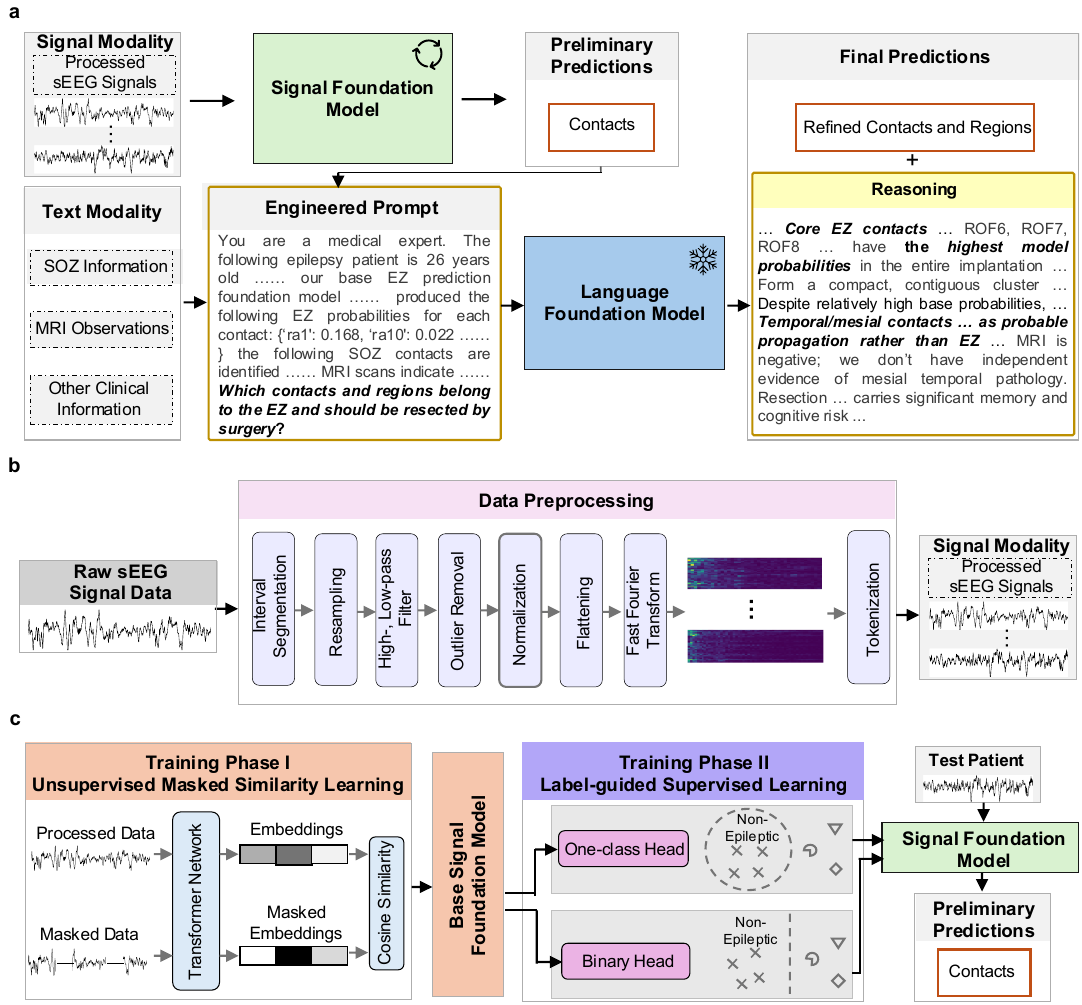}
    \caption{\small{\textbf{Overview of EpiiSLM.} \textbf{a}, End-to-end framework combining a trainable SFM for sEEG signals with a frozen LFM for integrating multimodal clinical priors and medical reasoning. The SFM encodes sEEG recordings to produce Preliminary EZ contact predictions and contact-level EZ probability scores. These outputs, together with structured clinical text (SOZ information, MRI findings and other clinical data) and an engineered prompt, condition the LFM to generate Final EZ predictions: Core and Possible Extended EZ contacts and regions, along with supporting medical reasoning. \textbf{b}, sEEG preprocessing pipeline for constructing signal-modality inputs to the SFM, comprising segmentation, resampling, filtering, artifact and outlier handling, normalization and flattening, FFT and tokenization. \textbf{c}, SFM training consists of two phases. Phase I uses unsupervised masked reconstruction with a similarity-learning framework on unlabelled sEEG recordings to learn general signal representations. Phase II applies label-guided supervised learning through two complementary heads (One-class and Binary), trained on resection extent, SOZ annotation and surgical outcome labels. At inference, a test patient's sEEG data passes through the SFM to produce the preliminary contact-level probabilities used in \textbf{a}. EZ, epileptogenic zone; FFT, fast Fourier transform; LFM, language foundation model; SFM, signal foundation model; sEEG, stereo-electroencephalography; SOZ, seizure onset zone.}}
\label{fig:model}
    \label{fig:model}
\end{figure}

\subsection{Each component of EpiiSLM contributes to EZ identification}\label{sec:ablation}

We conducted an ablation study to quantify the contribution of each proposed component in EpiiSLM, (Table~\ref{tab:compare_results}). To evaluate our SFM in isolation, we removed the LFM in EpiiSLM$_{(4)}$, which reduced PPV$_3$ by 4.0\% (-2.9\% PPV$_0$); yet performance remained 11.7\% above the SOZ-as-EZ baseline (+3.0\% PPV$_0$), indicating the standalone strength of our SFM. Removing Training Phase I in EpiiSLM$_{(3)}$ further reduced PPV$_3$ by 9.1\% (-9.6\% PPV$_0$) relative to EpiiSLM$_{(4)}$, underscoring the value of unsupervised representation learning on large-scale unlabeled sEEG data that prior work typically discarded. We then assessed the two-head ensemble: removing the Binary Head in EpiiSLM$_{(2)}$ decreased PPV$_3$ by 12.7\% (-13.6\% PPV$_0$), while removing the One-Class Head in EpiiSLM$_{(1)}$ decreased PPV$_3$ by 23.8\% (-25.5\% PPV$_0$) from EpiiSLM$_{(4)}$. These results confirm that ensembling two complementary heads, each trained with a non-ambiguous supervised learning scheme in Training Phase II, is essential to full model performance.

\renewcommand{\arraystretch}{1.22}
\setlength{\tabcolsep}{4.5pt}
\begin{table}[thb]
    \centering
    \scalebox{0.92}{
    \begin{tabular}{ccccccccccc}
    \hline
         \multirow{3}{*}{Method} & \makecell{SFM \\ Training Phase I} & \multicolumn{2}{c}{\makecell{SFM \\ Training Phase II}} & \multirow{3}{*}{LFM} & \multicolumn{3}{c}{PPV$_0$} & \multicolumn{3}{c}{PPV$_3$} \\ \cmidrule(lr){2-2} \cmidrule(lr){3-4} \cmidrule(lr){6-8} \cmidrule(lr){9-11}
         & Unsupervised & Binary & One-Class & & \makecell{Mean\\$\pm$ s.d.} & Median & \makecell{IQR \\(Q1-Q3)} & \makecell{Mean\\$\pm$ s.d.} & Median & \makecell{IQR \\(Q1-Q3)} \\\hline\hline
         SOZ-as-EZ & -- & -- & -- & -- & \makecell{0.718\\$\pm$0.260} & 0.667 & \makecell{0.500-\\1.000} & \makecell{0.827\\$\pm$0.220} & \underline{0.967} & \makecell{0.675-\\\textbf{1.000}} \\ \hline
         EpiiSLM$_{\text{(1)}}$ & \cmark & \cmark & -- & -- & \makecell{0.493\\$\pm$0.390} & 0.500 & \makecell{0.125-\\0.775} & \makecell{0.700\\$\pm$0.422} & \textbf{1.000} & \makecell{0.500-\\\textbf{1.000}} \\ \midrule
         EpiiSLM$_{\text{(2)}}$ & \cmark & -- & \cmark & -- & \makecell{0.612\\$\pm$0.440} & 0.804 & \makecell{0.150-\\\underline{0.977}} & \makecell{0.811\\$\pm$0.329} & \textbf{1.000} & \makecell{0.777-\\\textbf{1.000}}  \\ \midrule
         EpiiSLM$_{\text{(3)}}$ & --  & \cmark & \cmark & -- & \makecell{0.652\\$\pm$0.456} & \underline{0.861} & \makecell{0.200-\\\textbf{1.000}} & \makecell{0.847\\$\pm$0.319} & \textbf{1.000} & \makecell{0.850-\\\textbf{1.000}} \\ \midrule
         EpiiSLM$_{\text{(4)}}$\dag & \cmark & \cmark & \cmark & -- & \makecell{\underline{0.748}\\\underline{$\pm$0.366}} & 0.854 & \makecell{\textbf{0.797}-\\\textbf{1.000}} & \makecell{\underline{0.938}\\\underline{$\pm$0.112}} & \textbf{1.000} & \makecell{\underline{0.917}-\\\textbf{1.000}} \\ \midrule
         EpiiSLM\dag & \cmark & \cmark & \cmark & \cmark & \makecell{\textbf{0.777}\\\textbf{$\pm$0.353}} & \textbf{0.944} & \makecell{\underline{0.777}-\\\textbf{1.000}} & \makecell{\textbf{0.978}\\\textbf{$\pm$0.070}} & \textbf{1.000} & \makecell{\textbf{1.000}-\\\textbf{1.000}} \\ \hline         
    \end{tabular}}
    \caption{\textbf{EZ identification performance and ablation analysis of EpiiSLM.} The naïve baseline (row 1) simply takes SOZ as the EZ. Rows 2-5 report ablated variants: EpiiSLM$_{(1)}$ removes the One-Class Head and LFM; EpiiSLM$_{(2)}$ removes the Binary Head and LFM; EpiiSLM$_{(3)}$ removes Training Phase~I of the SFM and LFM; and EpiiSLM$_{(4)}$ removes only the LFM. The last row reports the full model. Checkmarks indicate included components; dashes indicate excluded components. We evaluated performance on Engel Class~Ia patients using PPV$_0$ and PPV$_3$, reporting mean, s.d., median, and IQR (Q1--Q3) to capture inter-patient variability. For EpiiSLM$_{(4)}$ and the full EpiiSLM, we averaged three independent runs before computing all reported statistics (marked by $\dag$). Default hyperparameters: BIOT and MedGemma backbones, all N3 data, all clinical priors included, the Transductive setting. \textbf{Bold} and \underline{underline} denote best and second-best values, respectively. EZ, epileptogenic zone; SOZ, seizure onset zone; SFM, signal foundation model; LFM, language foundation model; PPV, positive predictive value; s.d., standard deviation; IQR, interquartile range.}
    \label{tab:compare_results}
\end{table}

\subsection{Backbone selection improves signal and language model performance}\label{sec:backbones} 

\begin{figure}[t]
    \centering
    \includegraphics[width=1\linewidth]{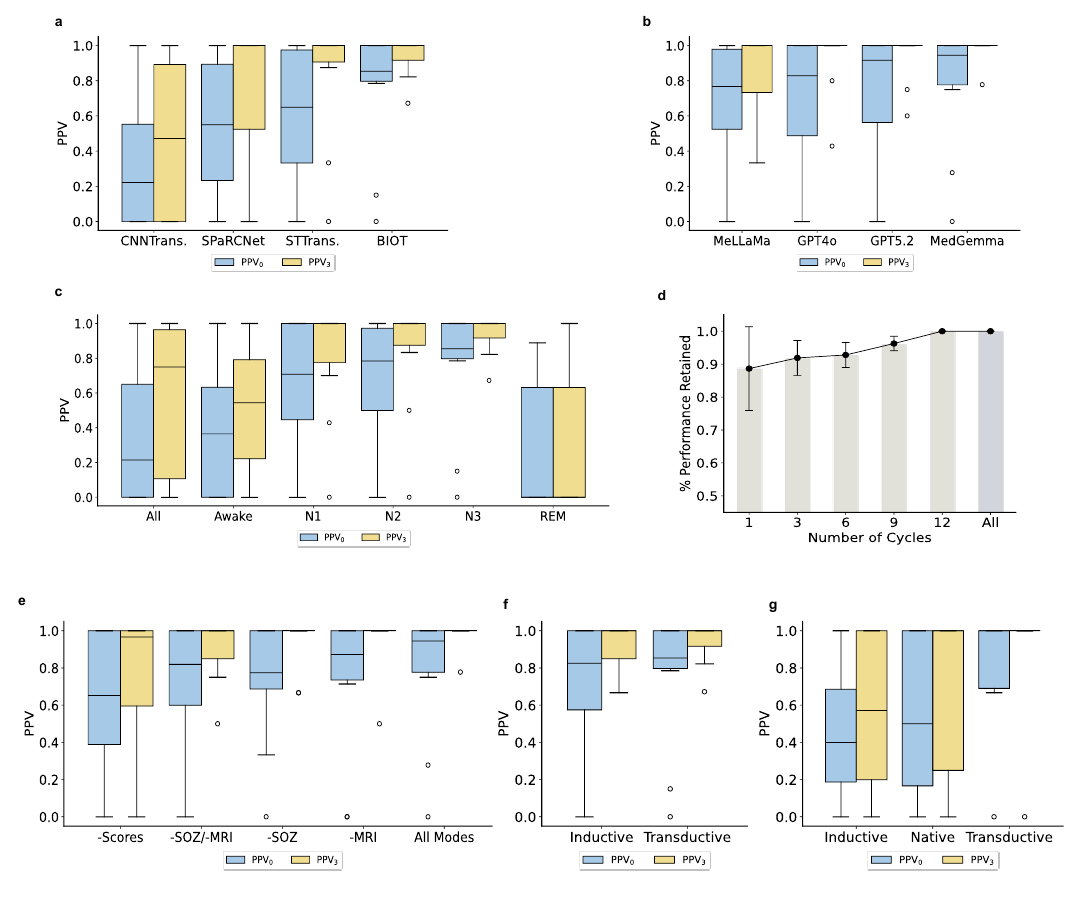}
    \caption{\small{\textbf{Extensive experimental evaluation of EpiiSLM.} \textbf{a}, SFM backbone comparison. \textbf{b}, LFM backbone comparison. \textbf{c}, Effect of vigilance state when only a single vigilance state is used during training phase II and inference. \textbf{d}, Test-time data efficiency: PPV retention as the number of N3 cycles used at inference is reduced. Error bars denote standard deviation across subsets of N3 data; `All' denotes each patient's full available N3 data. \textbf{e}, Effect of removing (denoted `-') individual clinical priors before the LFM. \textbf{f}, Generalization under internal patient-level distribution shifts. \textbf{g}, Generalization under external site-level distribution shifts. All experiments were evaluated using a leave-one-patient-out scheme with Engel class Ia patients; inter-patient variability is represented by standard box plots, where center lines indicate the median, boxes span from the first to the third quartile and whiskers extend to 1.5× the IQR. Unless otherwise stated, default EpiiSLM hyperparameters were used: BIOT and MedGemma backbones, all available N3 data, all clinical priors included, and the Transductive setting. \br{Corresponding per-patient results for \textbf{a}--\textbf{c},\textbf{e}--\textbf{g} are reported in Supplementary Note \b{7}.} SFM, signal foundation model; LFM, language foundation model; PPV, positive predictive value.}}
    \label{fig:all_fig_results}
\end{figure}

To select the SFM backbone best suited for EpiiSLM, we compared four biosignal foundation model architectures: BIOT \cite{biot}, STTransformer \cite{song2021transformer}, SPaRCNet \cite{jing2023development}, and CNNTransformer \cite{peh2022transformer}, using the signal-only variant EpiiSLM$_{(4)}$. We trained and evaluated all backbones under the same two-phase protocol and leave-one-patient-out scheme (Fig.~\ref{fig:all_fig_results}\textbf{a}). BIOT achieved the highest PPV$_3$ ($0.938\pm0.112$, median $=1.000$, IQR $=0.917$--$1.000$), followed by STTransformer ($0.821\pm0.356$, median $=1.000$, IQR $=0.906$--$1.000$), SPaRCNet ($0.710\pm0.418$, median $=1.000$, IQR $=0.525$--$1.000$) and CNNTransformer ($0.452\pm0.439$, median $=0.472$, IQR $=0.000$--$0.893$; $p=0.017$, Wilcoxon signed-rank test). We selected BIOT as the SFM backbone for all subsequent analyses on the basis of its superior accuracy and memory efficiency.

To select the LFM backbone for EpiiSLM, we evaluated four pretrained LFMs: MedGemma \cite{sellergren2025medgemma}, GPT-5.2 \cite{singh2025openai}, GPT-4o \cite{hurst2024gpt}, and MeLLaMA \cite{xie2025medical}, under the same leave-one-patient-out evaluation protocol, conditioning each on our SFM-derived contact-level EZ probabilities together with all available clinical priors (Fig.~\ref{fig:all_fig_results}\textbf{b}). MedGemma yielded the best zero-shot performance (PPV$_3$ = $0.978\pm0.070$, median $=1.000$, IQR $=1.000$--$1.000$), reflecting its specialized pretraining
on medical datasets \cite{sellergren2025medgemma}. GPT-5.2 followed ($0.935\pm0.142$, median $=1.000$, IQR $=1.000$--$1.000$), then GPT-4o ($0.923\pm0.185$, median $=1.000$, IQR $=1.000$--$1.000$) and MeLLaMA ($0.837\pm0.246$, median $=1.000$, IQR $=0.733$--$1.000$). We therefore adopted MedGemma as the default LFM in EpiiSLM. Qualitative examples of each LFM's outputs appear in Supplementary Note \b{12}.

\subsection{Deeper non-REM sleep improves EpiiSLM performance}\label{sec:sleep}

In Training Phase I, our SFM learns robust, general-purpose representations from a large and diverse sEEG corpus. In Training Phase II, it learns labeled class distributions from curated data, including data isolated to non-REM sleep rather than mixed vigilance states (Awake, N1, N2, N3, or REM). To examine how each vigilance state affects Training Phase II, we trained and evaluated the signal-only model (EpiiSLM$_{(4)}$) using data from individual states (Fig.~\ref{fig:all_fig_results}\textbf{c}).

Deeper non-REM sleep yielded progressively better performance. N3 data performed best (PPV$_3=0.938\pm0.112$, median $=1.000$, IQR $=0.917$--$1.000$), followed sequentially by N2 ($0.833\pm0.333$; median $=1.000$, IQR $=0.875$--$1.000$) and N1 (0.813 $\pm$ 0.344; median $=1.000$, IQR $=0.775$--$1.000$), Awake (0.507 $\pm$ 0.377; median $=0.544$, IQR $=0.221$--$0.792$) and REM (0.324 $\pm$ 0.437; median $=0.000$, IQR $=0.000$--$0.632$) data. \br{This trend is consistent with prior reports \cite{klimes2019nrem} that non-REM sleep facilitates the expression of biomarkers used for the EZ identification task, whereas the Awake and REM states are less informative due to increased high-frequency activity or physiological suppression.}\s{by N2 ($0.833\pm0.333$; median $=1.000$, IQR $=0.875$--$1.000$) and N1 (0.813 $\pm$ 0.344; median $=1.000$, IQR $=0.775$--$1.000$). This trend is consistent with prior reports that NREM sleep (stages N2 and N3) is the vigilance state that best identifies the EZ in sEEG data \cite{klimes2019nrem}. Performance dropped sharply for Awake (0.507 $\pm$ 0.377; median $=0.544$, IQR $=0.221$--$0.792$) and REM (0.324 $\pm$ 0.437; median $=0.000$, IQR $=0.000$--$0.632$) data, where increased artifacts from muscle movements and external stimuli degrade signal quality.} Both N3 and N2 data significantly outperformed all data combined (PPV$_3=0.579\pm0.436$, median $=0.750$, IQR $=0.107$--$0.964$; $p<0.05$, Wilcoxon signed-rank test), which is dominated by Awake data. We therefore use N3 data for all subsequent experiments when available.

\subsection{EpiiSLM achieves robust EZ identification with a single night's sleep}\label{sec:num_cycles}

Requiring only interictal N3 data at inference not only improves model performance but also raises the possibility of reducing the overall length of invasive sEEG recording. We tested EpiiSLM$_{(4)}$'s robustness to reduced inference data (Fig.~\ref{fig:all_fig_results}\textbf{d}). Following established sleep-cycle conventions \cite{altevogt2006sleep}, we partitioned each patient's N3 data into fixed 20-minute cycles (median = 12, max = 21 per patient). With only three N3 cycles--the lower-bound estimate of a single night's N3 sleep--EpiiSLM$_{(4)}$ retained on average 91.9\% of its full-data PPV$_3$ (95\% CI 67.8--100.0\%), with low variability across consecutive subsets (mean s.d. 5.3\%). These results indicate that EpiiSLM$_{(4)}$ is robust to both limited N3 data and varying data segments, unlike existing methods that require manual segment selection beyond sleep staging \cite{klimes2019nrem,nejedly2025leveraging}.

\subsection{EZ identification improves with multi-modal clinical priors}\label{sec:clinical_priors}

Even when ictal recordings are unavailable, the signal-only system EpiiSLM$_{(4)}$ produces accurate EZ predictions from interictal data alone (Table~\ref{tab:compare_results}). When seizures are captured and SOZ labels can be derived, or when MRI observations \cite{spanedda1997relations} are available, the full EpiiSLM integrates these priors to further improve performance. Fig.~\ref{fig:all_fig_results}\textbf{e} quantifies the contribution of each source. Using all inputs yielded a PPV$_3$ of $0.978\pm0.070$ (median $=1.000$, IQR $=1.000$--$1.000$). Removing MRI reduced performance to $0.950\pm0.158$ (median $=1.000$, IQR $=1.000$--$1.000$); removing SOZ, to $0.933\pm0.141$ (median $=1.000$, IQR $=1.000$--$1.000$); and removing both, to $0.905\pm0.171$ (median $=1.000$, IQR $=0.850$--$1.000$). Conversely, when we removed the SFM-derived contact probabilities and relied only on clinical priors, PPV$_3$ dropped to $0.774\pm0.329$ (median $=0.967$, IQR $=0.595$--$1.000$)--a 20.4\% reduction relative to the full system ($p<0.05$; Wilcoxon signed-rank test), confirming that the learned sEEG features are the dominant driver in EpiiSLM's performance.

\subsection{EpiiSLM generalizes across patients and hospital sites}\label{sec:generalization}

A key clinical requirement is robustness to distribution shifts arising from inter-patient variability, recording conditions (electrode type, implantation geometry), and differences in data-acquisition systems across hospitals. We designed EpiiSLM's training protocol to address these shifts through two elements: the large MNI prior dataset and a transductive learning mechanism. Because Training Phase I is unsupervised, we can include a new patient's unlabeled recordings to adapt the signal representation space to that patient's distribution, while Training Phase II uses only prior patients with known resection labels and surgical outcomes. This allows the model to incorporate target-distribution knowledge into its sEEG representations without any information about the test patient's SOZ, resection, or outcome.

\paragraph{Internal patient-level distribution shifts}
Within the MNI cohort, distribution shifts reflect inter-patient variability and changes in acquisition protocols. We evaluated EpiiSLM$_{(4)}$ (SFM only) under (i) an Inductive setting, where the held-out patient's data is excluded from Training Phase I, and (ii) a Transductive setting, where unlabeled test-patient data is included (Fig.~\ref{fig:all_fig_results}\textbf{f}). Transductive learning improved PPV$_3$ from $0.927\pm0.124$ (median $=1.000$, IQR $=0.850$--$1.000$) to $0.938\pm0.112$ (median $=1.000$, IQR $=0.917$-$1.000$), indicating that EpiiSLM(4) already performs well without domain adaptation but benefits from it nonetheless. In a separate analysis restricted to commercial-electrode recordings ($n=7$, excluding homemade electrodes), EpiiSLM$_{(4)}$ achieved a perfect PPV$_3$ of $1.000\pm0.000$ (median $=1.000$, IQR $=1.000$--$1.000$) and a PPV$_0$ of $0.932\pm0.084$ (median $=1.000$, IQR $=0.873$--$1.000$); full details appear in Supplementary Note \b{2}.\s{MOVED TO DISCUSSION: In practice, the Inductive setting with a fixed pretrained model enables immediate EZ predictions, whereas the Transductive setting requires an additional training run (typically $\sim$1–2 days depending on hardware)—a minor cost within the surgical planning window (weeks to months) that is justified by the increased likelihood of seizure freedom and functional preservation.}

\paragraph{External site-level distribution shifts}

We evaluated cross-site generalization from MNI to an independent dataset from St Anne's University Hospital (Brno, Czech Republic) comprising 17 patients, 7 of whom were Engel Class Ia. The Brno distribution differs substantially from MNI: recordings are approximately $14\times$ shorter for training and $4\times$ shorter for inference, the initial sampling rate is higher (5,000 Hz), and further differences exist in electrode type, line-noise frequency (50 Hz versus 60 Hz), power spectral density, interval-level statistics, and recording environment (Supplementary Note \b{10}). Because N3 data were unavailable for all Brno Engel Class Ia patients, we used available N2 data for Training Phase II and inference, which also yielded improvement over using all sleep stages (see Section~\ref{sec:sleep}).
 
We compared EpiiSLM$_{(4)}$ on the Brno Engel Class Ia patients under three settings using one-patient-out evaluation: (i) Inductive (MNI data only), (ii) Native (Brno data only), and (iii) Transductive (MNI + Brno data) (Fig.~\ref{fig:all_fig_results}\textbf{g}). Despite the substantial site-level differences and using N2 over N3, EpiiSLM$_{(4)}$ achieved an Inductive PPV$_3$ of $0.567\pm0.420$ (median $=0.571$, IQR $=0.200$--$1.000$), quantifying its inductive generalizability without any target-hospital knowledge. Training directly on Brno data improved Native PPV$_3$ to $0.643\pm0.440$ (median $=1.000$, IQR $=0.250$--$1.000$), but the small dataset size hindered full-distribution learning. When we combined source and target hospital data, Transductive PPV$_3$ reached 0.857 (95\% CI 0.429--1.000, median $=1.000$, IQR $=1.000$--$1.000$), showing that our adaptation improves generalization under site-level distribution shifts. 

\subsection{Qualitative evaluation confirms alignment between EpiiSLM predictions and surgical outcomes} \label{sec:qualitative}

\begin{figure}[!ht]
    \centering
    \begin{minipage}{0.49\textwidth}
        \centering
        \includegraphics[width=\linewidth]{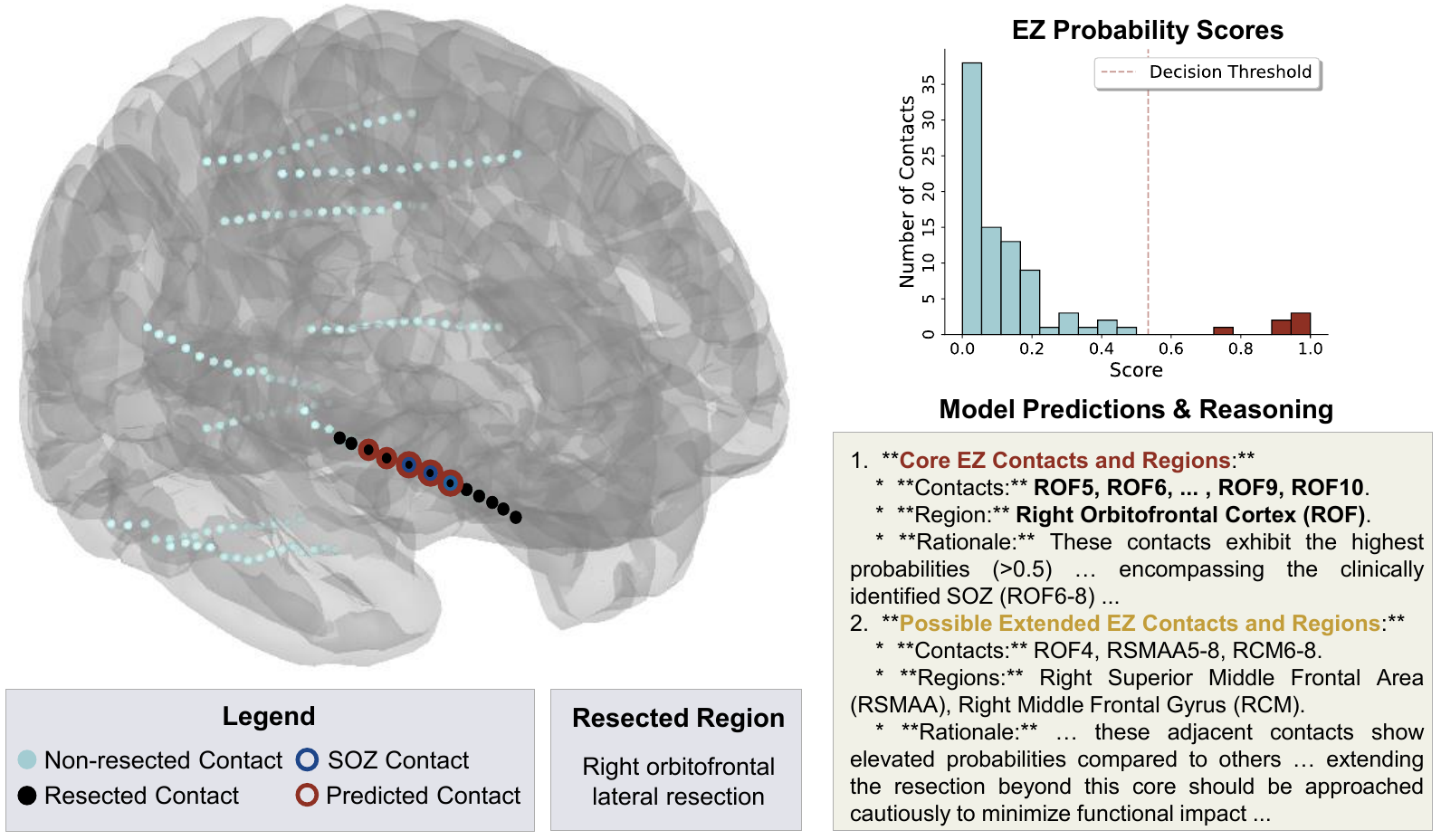}
        \footnotesize{Patient 1 (Ia)}
    \end{minipage}
    \hfill
    \begin{minipage}{0.49\textwidth}
        \centering
        \includegraphics[width=\linewidth]{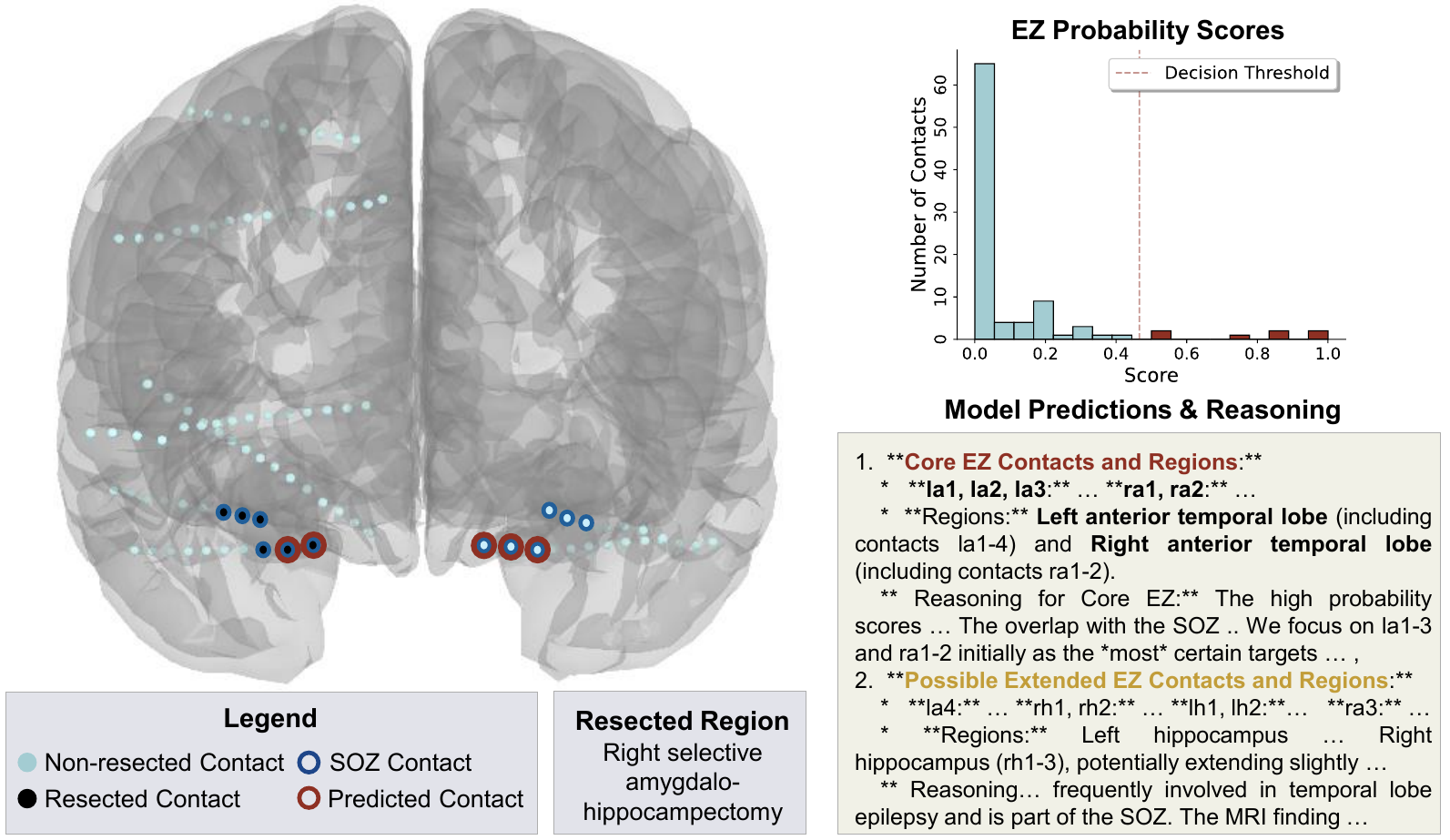}
        \footnotesize{Patient 13 (Ib)}
    \end{minipage}
    \hfill
    \begin{minipage}{0.49\textwidth}
        \centering
        \includegraphics[width=\linewidth]{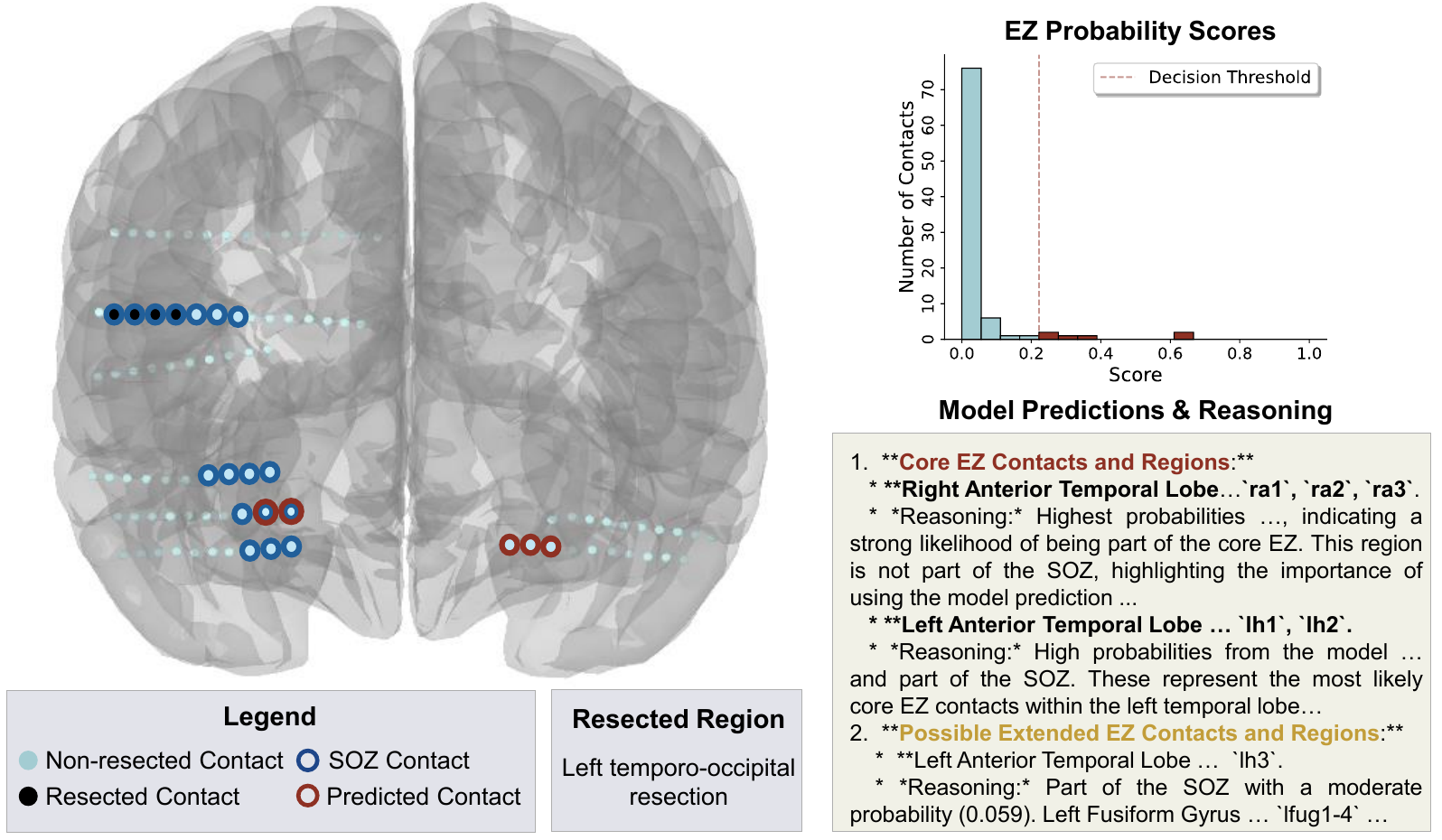}
        \footnotesize{Patient 18 (IIIa)}
    \end{minipage}
    \hfill
    \begin{minipage}{0.49\textwidth}
        \centering
        \includegraphics[width=\linewidth]{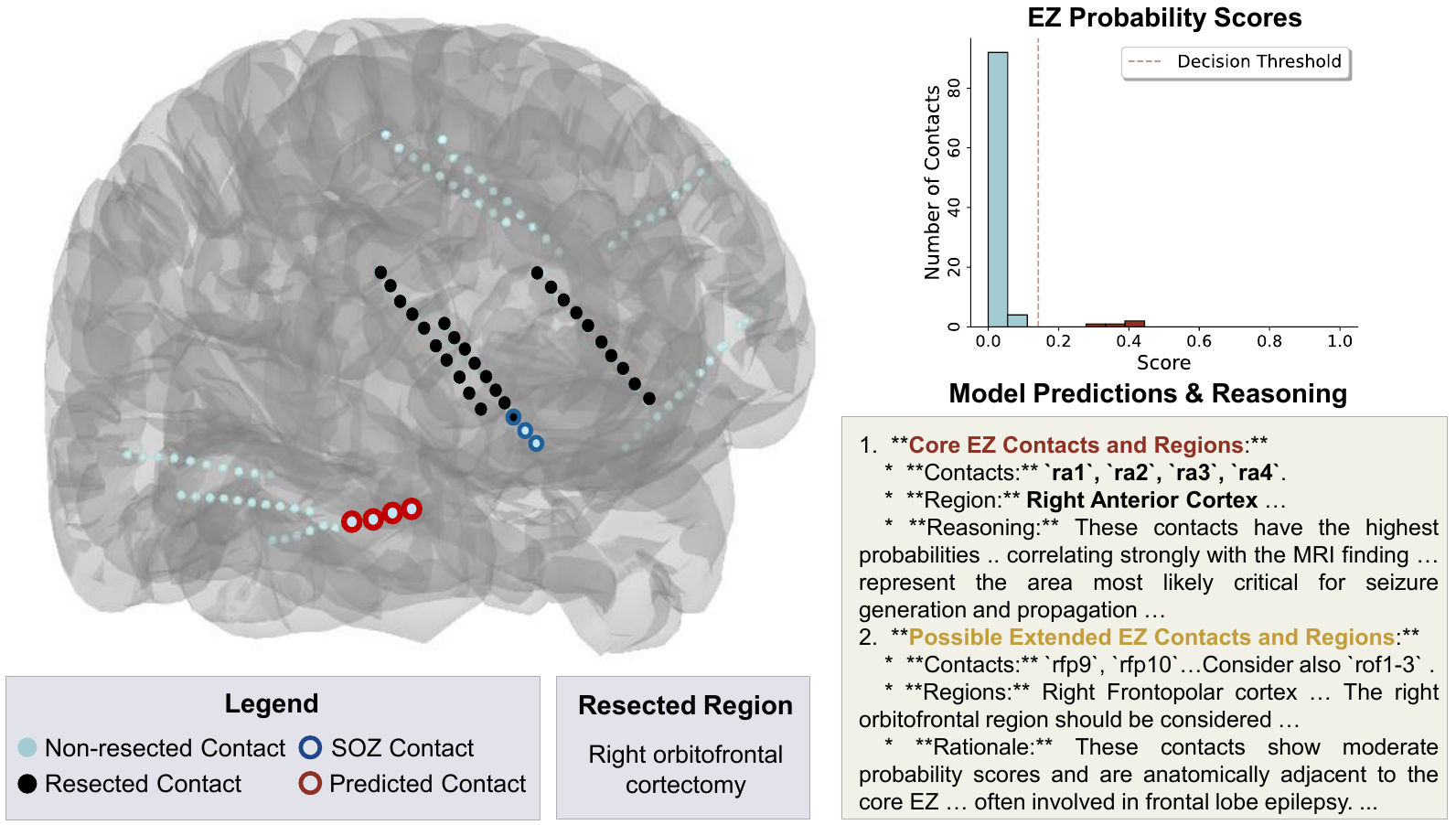}
        \footnotesize{Patient 26 (IVa)}
    \end{minipage}
    \caption{\small{\textbf{Qualitative EpiiSLM predictions and surgical outcomes for four representative patients.}\s{\textbf{a}, Engel Class Ia. \textbf{b}, Engel Class Ib. \textbf{c}, Engel Class IIIa. \textbf{d}, Engel Class IVa.} Each panel shows the distribution of contact-level EZ probability scores with the 95th-percentile threshold defining the high-confidence set, a brain map of contact-level EZ predictions overlaid with surgical ground truth (for clarity, a subset of sEEG contacts is displayed), the surgically resected region, and an excerpt of EpiiSLM's clinical report including contact- and region-level predictions with medical reasoning. Predicted contacts and regions are highlighted. sEEG, stereo-electroencephalography; EZ, epileptogenic zone.}}
    \label{fig:visualization_llm}
\end{figure}

We visualized EpiiSLM predictions alongside surgical ground truth and outcomes for four representative patients spanning Engel Class Ia, Ib, IIIa, and IVa outcomes in Fig.~\ref{fig:visualization_llm} (Engel Class definitions in Section~\ref{subsec:datasets}; all-patient visualizations in Supplementary Notes \b{13--14}). For the Engel Class Ia case, we detected EZ contacts within the surgically resected region, though not all resected contacts were predicted as EZ, consistent with the principle that removing key EZ contacts, rather than the entire resection volume, drives seizure freedom. For the Engel Class Ib case, EpiiSLM identified additional EZ contacts beyond the resection boundary, consistent with the patient's continued but reduced post-surgical seizures. For the Engel Class IIIa and IVa cases, high-confidence EZ contacts lay entirely outside the resection, suggesting that the resected volume did not effectively capture the epileptogenic network, consistent with the patients' poor surgical outcomes. \s{We could remove this for NM because it's already in discussion and based on Jean's comment, it seems like this is too obvious: Together, these cases underscore the importance of precise EZ localization for surgical planning and reinforce that SOZ and resection margins are imperfect proxies for the true EZ.}

We present an excerpt of EpiiSLM's clinical report for each patient alongside the brain visualizations in Fig.~\ref{fig:visualization_llm}. At the region level, EpiiSLM achieved 100\% accuracy among Engel Class Ia patients, as the predicted and resected regions in Fig.~\ref{fig:visualization_llm} show. The report further provides stratified Core EZ and Possible Extended EZ categories, EZ probability scores for every contact, and medical reasoning citing overlap or mismatch between high-scoring contacts, SOZ, and MRI lesion areas. Together, these outputs allow neurophysiologists to incorporate their own judgment, weigh seizure control against functional risk, and plan surgery informed by the full prediction context.

\section{Discussion}\label{discussion}

EpiiSLM addresses the two persistent bottlenecks in sEEG-based EZ identification (inefficient data use and ambiguous labels) through a dual foundation model design that leverages typically discarded poor-outcome data through unsupervised representation learning, anchors supervised learning on the well-defined non-epileptic class,  and avoids handcrafted features and manually selected data segments unlike previous methods \cite{klimes2019nrem,nejedly2025leveraging} with end-to-end learned features. EpiiSLM achieved 100\% region-level accuracy and high contact-level precision: 0.777 PPV$_0$ and 0.978 PPV$_3$, significantly outperforming the SOZ-as-EZ baseline ($p<0.05$). Although direct comparison across studies is infeasible (Section \ref{sec:general_performance}), these results appear as a substantial advance over reported PPV$_0$ values in the literature, which do not exceed 64\% \cite{klimes2019nrem, makaram2023deep, balaji2024seizure, nejedly2025leveraging, klimes2024interictal}.

Cross-site evaluation on an external cohort from St Anne's University Hospital (Brno) showed that our transductive adaptation mechanisms raised PPV$_3$ from 0.567 to 0.857, showing that the EpiiSLM framework itself can generalize to an external site, and that the large MNI corpus can support domain adaptation when target distribution data is limited. In practice, the Inductive setting yields immediate EZ predictions, whereas the Transductive setting would require an additional training run ($\sim$1–2 days depending on hardware) using only data that is already available in this context; a minor cost within the surgical planning window of weeks to months, justified by improved model performance.

Vigilance-state analyses revealed that N3 sleep is the most informative state for interictal EZ identification, consistent with prior evidence \cite{klimes2019nrem}. EpiiSLM$_{(4)}$ requires no ictal data and retains 91.9\% of its full-data PPV$_3$ with as few as three N3 cycles ($\sim$one night of sleep), while traditional sEEG monitoring averages 12.7 days \cite{klimes2019nrem} because neurologists must wait to capture ictal events to localize the SOZ. EpiiSLM could therefore substantially shorten this invasive monitoring period and reduce associated procedural risk. If seizures do occur during the recording window, the full EpiiSLM framework incorporates SOZ information to refine its prediction.

Within the full EpiiSLM framework, removing SOZ and MRI priors reduced PPV$_3$ by only 7.3\%, whereas removing the SFM-derived contact probabilities while retaining SOZ and MRI priors reduced PPV$_3$ by 20.4\% ($p<0.05$), showing that conventional priors used in surgical planning are thus informative but insufficient on their own, whereas the sEEG signal features learned by our foundation model are essential for accurate EZ identification. The signal-only variant EpiiSLM$_{(4)}$ is therefore recommended when clinical priors are incomplete. When SOZ and MRI priors are available, however, the full EpiiSLM not only improves precision but also provides medical reasoning that can serve as a basis for surgical planning discussions.

We designed EpiiSLM to assist, not replace, clinical decision-making during surgical planning. The system produces per-contact EZ probability scores with a tunable precision-recall trade-off: clinicians can prioritize the most probable contacts to minimize unnecessary resection, or broaden the set when maximizing seizure freedom is the primary goal. Because post-resective seizure freedom currently remains below 50\% \cite{malmgren2017long}, even modest gains in EZ-identification accuracy could benefit the millions of patients with drug-resistant epilepsy \cite{surges2021identifying}. More precise localization also reduces the risk of resecting healthy brain tissue--a critical consideration given the cognitive, sensory, and motor deficits that over-resection can cause.

An intrinsic constraint of sEEG-based EZ identification is small cohort size, which is due to the invasive nature of sEEG \cite{mullin2016seeg}, low seizure-freedom rates \cite{Avigdorjnnp-2025-337158,malmgren2017long}, and privacy and standardization limitations in cross-site data sharing and acquisition protocols \cite{rahimzadeh2023benefits}. While our evaluation with 17 Engel Ia patients (47 Engel I-IV patients for training) across two sites may validate the principles of EpiiSLM, larger multicenter cohorts could further establish the generalizability of exact EpiiSLM configurations. Cohort size also limits formal fairness inference: subgroup performance was consistent for the primary PPV$_3$ metric, but sex and electrode type were confounded (all homemade-electrode patients were female), and the principal failure mode was reduced precision on non-commercial homemade electrodes (Supplementary Notes \b{2--4}). We also excluded formal quantitative evaluation of EpiiSLM's reasonings, which is still an open challenge as current methods rely on costly, noisy expert assessment \cite{savage2024diagnostic}\s{lightman2023let,}. Finally, EpiiSLM depends on experts to identify MRI lesions and determine the SOZ from ictal sEEG; though more straightforward than EZ identification, automating these with text-based outputs is a natural next step, as is pairing EpiiSLM's natural-language interpretability with explainable-AI methods to explicitly identify sEEG biomarkers it currently learns implicitly.

In summary, our results validate EpiiSLM's dual-foundation model design, large-scale unsupervised sEEG learning, and grounded non-epileptic class approach for improving EZ-identification accuracy, which has the potential to increase seizure-freedom rates after resective surgery for millions of patients with drug-resistant epilepsy. Producing predictions from a single night of interictal sEEG could further shorten invasive monitoring and reduce procedural risk. Future studies should pursue clinical validation toward regulatory approval and integration into surgical planning protocols.

\section{Methods}\label{methods}

\subsection{Data curation} \label{subsec:datasets}

Our primary dataset comprised 30 patients (17 female / 13 male) with drug-resistant epilepsy from the Montreal Neurological Institute \& Hospital (MNI) database (Montreal, Quebec, Canada). All patients underwent sEEG monitoring followed by resective surgery between 2010 and 2017, with a minimum post-surgical follow-up of one year. Each patient had at least 40 hours of continuous sEEG recording, including at least one full night of sleep, yielding 104,990 minutes of sEEG data in total. The cohort spans a range of ages, pathologies, and regions of interest (Table \ref{tab:clinical-characteristics}). For external evaluation, we additionally used sEEG data from 17 patients at St Anne's University Hospital (Brno, Czech Republic), totalling 7,106 minutes of recordings. The full study cohort therefore comprised 47 patients across the two sites.

We classified post-surgical outcomes using the Engel system \cite{engel1993update}. Class I (free of disabling seizures) is subdivided into Ia (completely seizure-free since surgery), Ib (non-disabling simple partial seizures only), Ic (free of disabling seizures for at least two years after some post-surgical disabling seizures), and Id (generalized convulsions only on antiepileptic-drug withdrawal); Classes II--IV correspond to progressively worse outcomes, from rare disabling seizures (Class II) to persistent disabling seizure (Class III) and no worthwhile improvement (Class IV). Of the 30 MNI patients, 10 (33.3\%) had Engel Ia outcomes and 20 (66.7\%) had outcomes ranging from Ib to IVb. Although patients classified as Engel Ib–Id technically fall under Class I, they continued to experience seizures post-surgery and were not completely seizure-free, hence, their outcome labels are less reliable for model evaluation purposes. We therefore performed quantitative model evaluation on the 10 Engel Ia patients; the remaining 20 patients contributed to model training only.


For the MNI cohort, \br{inclusion criteria followed \cite{klimes2022spatio}}. sEEG signals were recorded with the Harmonie EEG system (Stellate, Montreal, Canada) at a sampling rate of 2000 Hz, with a low-pass filter at 600 Hz and a high-pass filter at 0.3 Hz. For the Brno cohort, signals were recorded with the BrainScope system (M\&I, BrainScope, Czech Republic) at 5000 Hz with a low-pass filter at 2000 Hz; a high-pass filter at 0.3 Hz was applied post-acquisition. Two neurologists identified SOZ contacts by consensus, defined as the contacts showing the earliest electrographic changes at seizure onset across all recorded clinical seizures \cite{spanedda1997relations}. We identified resected contacts by co-registering post-resection imaging with pre-surgical sEEG electrode locations using the MNI Open iEEG Atlas \cite{frauscher2018atlas}. Moreover, we performed sleep staging in 30-second epochs from scalp EEG, electrooculography (EOG), and electromyography (EMG) signals, following the American Academy of Sleep Medicine (AASM) standard \cite{berry2012rules}.

\subsection{Data preprocessing}

We first extracted interictal and ictal segments from the long-duration sEEG recordings and divided both into 30-second time intervals. Ictal intervals, including a 1-hour buffer before and after each seizure, were discarded. Each interval contained the signal of a single electrode contact. We downsampled all signals to 200 Hz from the original 2000 Hz (MNI) and 5000 Hz (Brno).

To mitigate artifacts from muscle movement and other sources, we applied a patient-specific outlier-removal procedure. For each patient, we computed a set of interval-level statistics: mean, median, standard deviation, maximum amplitude, robust $Z$-score, kurtosis, bandpower in eight frequency bands (1--4, 4--8, 8--12, 12--20, 20--45, 45--65, 65--80, and 80--100 Hz), and line-noise peak power (60~Hz). We set patient-specific thresholds at the 90th percentile for peak-power features and the 99th percentile for all other features. Threshold values are reported in Supplementary Note \b{10}. An interval was flagged as an outlier if \emph{any} of its features exceeded the corresponding threshold. The same procedure was applied to both training (Phase II) and inference data, with thresholds computed independently for each patient.

\subsection{Overview of model design and training}

EpiiSLM is a dual foundation model system combining a signal foundation model, trained for this study, with a frozen language foundation model pretrained on medical question-answering data. As shown in Fig.~\ref{fig:model}, the SFM processes sEEG recordings and outputs Preliminary EZ contact predictions with contact-level EZ probability scores. The LFM receives these outputs together with multimodal clinical priors, including SOZ annotations, MRI-derived observations, and additional clinical information, through a structured prompt and produces the Final EZ predictions (Core EZ Contacts, Core EZ Regions, Possible Extended EZ Contacts, and Possible Extended EZ Regions) with supporting medical reasoning.

\subsection{Signal foundation model}

We adopted a transformer-based architecture for the signal foundation model due to its compatibility with BERT-style \cite{kenton2019bert} unsupervised pretraining, which enables the model to learn generalizable representations from large-scale unlabeled sEEG recordings. Transformer architectures, originally developed for natural language processing, have been widely adapted for signal processing, where they outperform recurrent \cite{hojjati2023multivariate} and graph neural networks \cite{ho2023graph} while better capturing long-range temporal dependencies \cite{biot}. We evaluated four high-performing candidate backbones: BIOT \cite{biot}, SPaRCNet \cite{jing2023development}, STTransformer \cite{song2021transformer}, and CNNTransformer \cite{peh2022transformer}. Among these, we selected BIOT as the default backbone for its ability to handle variable interval lengths, varying numbers of contacts, and different sampling rates, and for its strong performance both in our experiments (see Section~ \ref{sec:backbones}) and in previously published reports.

\subsubsection{Encoder architecture}

The sEEG signals in our datasets spanned a broad amplitude range. We therefore normalized each interval independently using the 95th percentile of the absolute amplitude within that interval \cite{biot}, rather than normalizing all contacts jointly as in previous work \cite{tang2021self, ho2023self}. We partitioned each normalized interval into overlapping fixed-length tokens with a hop length of 100 samples. To preserve complementary time- and frequency-domain information, we applied the Fast Fourier Transform (FFT)  to each token and concatenated the resulting frequency-domain representation with the raw time-domain signal. A learnable multi-layer perceptron (MLP) then projected the concatenated features into token embeddings. We added sinusoidal positional embeddings \cite{vaswani2017attention} to encode sequential order. The final input embedding for each token was obtained by summing its segment embedding with the corresponding positional embedding.
The token sequence was then processed by a linear-transformer block \cite{katharopoulos2020transformers}, which scales linearly with interval length rather than quadratically \cite{katharopoulos2020transformers, biot}, reducing memory requirements for the relatively large number of tokens per interval in our task. The block outputs interval-level representations for downstream tasks. A description of each architectural layer is provided in the Implementation Details.

\subsubsection{Training Phase I}

Training Phase I followed the unsupervised (or self-supervised) pretraining paradigm of BERT-like language models. We trained the signal foundation model encoder on our full sEEG corpus without any EZ-related labels. The goal was to learn general, lower-dimensional representations of interictal sEEG intervals that transfer to EZ identification in Training Phase II. Because no surgical ground truth was required, this phase included recordings from all patients, including 20 patients with poor surgical outcomes. Although these patients' data cannot support EZ evaluation since their EZ cannot be reliably inferred from the resection, their sEEG recordings contain generalizable signal dynamics that improve representation quality. We included intervals from all vigilance states and did not apply outlier removal in this phase; this broader data exposure improved generalizability and downstream performance in our experiments (see Section~\ref{sec:generalization}).
To train the model, we used an unsupervised masked reconstruction objective requiring no task-specific labels, following the protocol described in \cite{biot}. Each sEEG interval was perturbed by randomly dropping a subset of its electrode contacts and tokens before being passed through the encoder and a reconstruction head. The same interval was simultaneously processed without perturbation. The model was trained with a contrastive loss \cite{he2020momentum} that maximized the similarity between the two outputs, teaching the model to recover unperturbed representations from its perturbed counterpart \cite{pan2023towards}.

\paragraph{Distribution-shift mitigation}
We considered two settings for Training Phase I. In the Inductive setting, the encoder was pretrained on recordings from poor-outcome patients only. \br{Hence, data leakage is strictly prevented.} In the Transductive setting, Training Phase I was re-run with the held-out test patient's unlabeled sEEG data included alongside all available training patients regardless of surgical outcome. This adapted the representation space to the test patient's signal distribution. The procedure is clinically feasible because Training Phase I requires no labels (no SOZ, resection, or outcome information from the test patient is used), and the additional computation falls within the presurgical planning timeline. The empirical benefit of Transductive adaptation is reported in Section~\ref{sec:generalization}.

\subsubsection{Training Phase II}
Training Phase II was the supervised fine-tuning stage, in which the pretrained encoder was adapted for EZ identification through label-guided learning. Two copies of the Training Phase I encoder were fine-tuned separately as the One-Class Head and the Binary Head. Because contact-level EZ labels are not directly observable, we curated a supervised training set using only labels that can be reliably assigned at the contact level, i.e., SOZ and resected contacts. Both heads were grounded in the non-epileptic class, defined as signals from contacts that were neither resected nor annotated as SOZ in patients who were seizure-free (Engel Ia) after surgery. Moreover, both heads used SOZ contacts, which serve as positive counterexamples to sharpen the non-epileptic decision boundary. To represent the non-epileptic class as cleanly as possible, non-resected contacts annotated as SOZ were excluded from this class. Technically,
these contacts are by definition outside of the EZ since they were not resected but the patients have obtained seizure-freedom. However, many of these contacts exhibit overlap in characteristics with EZ contacts, and most of them fall within the margin of co-registration or electrode sensitivity error, which is consistent with our observations in Supplementary Notes \b{5,8--9}. We therefore treated them as generally epileptic. In an ideal scenario, if the amount of non-EZ SOZ data is sufficient, the negative class could be refined to include SOZ contacts that are epileptic but fall outside the true EZ.
Lastly, resected contacts were also excluded from supervised training because a subset of resected tissue may be non-epileptic, and assigning them outside the non-epileptic class would not be correct. 

Whereas Training Phase I prioritized data diversity and generalizability, Training Phase II prioritized a tight, clean data distribution. We filtered Phase II data by vigilance state and retained only non-outlier intervals. Vigilance-stratified performance is reported in the main paper, and outlier-inclusive performance is presented in Supplementary Note \b{5}. We further restricted Phase II to patients with Engel Class I outcomes (Ia–Id). Only Engel Class Ia patients are completely seizure-free, so we used them for evaluation; we nevertheless included Engel Ib–Id patients in training to enlarge and diversify both the non-epileptic and SOZ examples. Specifically, Ib–d patients contributed approximately one-third of the supervised training set, which empirically improved our model's learned distributions of the non-epileptic and epileptic SOZ subclass. However, including non-epileptic data from Ib–d patients introduces a risk of class contamination because a small fraction of contacts labeled non-epileptic could in fact be non-resected EZ. We first mitigated this risk by excluding all resected and SOZ contacts from the non-epileptic class. Additionally, in our Ib–d cases, high-risk contacts were most often located at extended resection margins or were unresected SOZ contacts spared for functional considerations, consistent with prior reports that postoperative seizure persistence can arise from incomplete resection near margins or from additional, unresected foci \cite{surges2013reoperation,englot2014factors,andrews2019association}. Ib–d patients also constitute a more favorable postoperative subgroup than Engel Class II–IV, with correspondingly lower likelihood of substantial unresected epileptogenic tissue \cite{surges2013reoperation}. The high class imbalance in sEEG datasets (non-epileptic $>>$  EZ) further attenuates the residual risk. Empirically, the remaining label ambiguity in these near-seizure-free patients was outweighed by the gain in supervised signal, as shown in Supplementary Note \b{5}.

As mentioned earlier, evaluation followed a leave-one-patient-out protocol. The MNI cohort contained 15 Engel Class I patients, of whom 10 were Engel Class Ia (completely seizure-free). In each fold, we performed Training Phase II on the remaining 14 Class I patients (including non-Ia) and evaluated the model on one held-out Class Ia patient. We repeated this for each of the 10 Class Ia patients and reported the mean $\pm$ s.d. across folds. This protocol ensured strict patient-level separation between training and test sets in every fold.

\paragraph{One-Class Head}

The One-Class head learns a compact representation of the \emph{non-epileptic} class (Class 0), defined as contacts that were neither SOZ nor resected, by mapping Class 0 examples to a hypersphere in the embedding space, using the Deep Support Vector Data Description (Deep SVDD) objective \cite{Ruff2020Deep}. Starting from the foundation model produced by Training Phase I, we initialized the hypersphere center as the mean embedding from an initial forward pass over Class 0 examples and then optimized the head under the Deep SVDD objective using these examples only. At inference, the head scored each example by its squared Euclidean distance to the hypersphere center. Small distances indicate non-epileptic–like activity, whereas large distances flag anomalous patterns treated as potential EZ candidates.

In practice, interictal sEEG signals from non-epileptic and potential EZ contacts differ only subtly, and a tight non-epileptic hypersphere alone may not reliably separate the two. We therefore augmented the Deep SVDD objective with a repulsion term \cite{Ruff2020Deep} that explicitly pushes non-Class 0 examples away from the hypersphere center. As counterexamples, we used SOZ-labeled contacts as the most reliable contact-level label of epileptic activity. This formulation is analogous to the open-set anomaly-detection setting \cite{lai2023open}, in which the model learns the distribution of normal data (here, non-epileptic activity) while seeing labeled examples of one anomaly type (here, SOZ contacts) during training, enabling detection of both seen and unseen anomalies at test time. Because the One-Class score is defined solely by distance to the non-epileptic center, the head's EZ predictions are determined only in relation to the learned non-epileptic distribution. By this design, the head can flag potential EZ contacts beyond those resembling SOZ patterns seen during training.

\paragraph{Binary Head}

The Binary head separates Class 0 (non-epileptic) from Class 1 (non-Class 0, comprising SOZ contacts). Unlike the One-Class head, which models only the Class 0 distribution, the Binary head learns an explicit decision boundary between the two classes. Architecturally, the Binary head consists of a fully connected layer that maps interval embeddings from the base signal foundation model to a single logit, with sigmoid output for binary classification. As in the One-Class head, the learning objective is anchored on the non-epileptic class, with SOZ contacts providing the non-Class 0 (Class 1) samples. This framing differs from previous EZ-identification approaches that pose the task as binary classification between a surrogate ``EZ'' class (resected contacts \cite{makaram2023deep,nejedly2025leveraging}, SOZ contacts \cite{luders2006epileptogenic}, or the intersection of the two \cite{klimes2019nrem}), and all remaining contacts. Such methods, in effect, learn to perform resected or SOZ prediction rather than EZ prediction. Resected contacts can include tissue that is not part of the EZ, and the non-SOZ class can include EZ contacts because the EZ commonly extends beyond the SOZ \cite{luders2006epileptogenic}. Both classes in those formulations are therefore label-ambiguous with respect to the EZ. Our Binary head sidesteps this ambiguity by anchoring on the non-epileptic class, which is operationally defined and unambiguous in the EZ-identification context, and using SOZ contacts as Class 1, which are by construction disjoint from Class 0. We accordingly interpret the binary score as a measure of non-epileptic resemblance, especially when combined with the One-Class head. To address the extreme class imbalance, we optimized the head with class-balanced binary cross-entropy \cite{cui2019class}.

\subsubsection{Preliminary EZ contact-level predictions}\label{sec:prelim}

At inference, the Binary head outputs a probability in $[0,1]$ for each interval, with values near 0 indicating the non-epileptic class and values near 1 indicating Class 1 (the SOZ class). The One-Class head produces a nonnegative score in $[0,\infty)$ (the squared Euclidean distance from the interval embedding to the hypersphere center), where scores near 0 indicate proximity to the non-epileptic distribution and larger values indicate deviation from it. We then applied the first of two thresholds to each interval to assign an interval-level prediction. Specifically, outputs exceeding this threshold received label 1 (potential EZ), and the remainder received label 0. The threshold is user-tunable along the precision–recall trade-off; we set it to 0.9 to prioritize precision over recall, on the basis that false positives risk contributing to over-resection of non-epileptic tissue. For each head, we aggregated interval-level scores into a contact-level score by averaging across all intervals at that contact. The final contact-level EZ probability was the average of the two heads' contact-level scores.

To obtain Preliminary EZ contact predictions, we applied a second adaptive threshold to each patient's distribution of contact-level EZ probability scores. Because contact-level score distributions vary substantially across patients, we set the threshold to the patient's own 95th percentile rather than a fixed global value. This retains the top 5\% of contacts per patient, a cutoff strict enough to suppress background scores but permissive enough to include more than the most extreme outliers. We then refined the above-cutoff contacts with two spatial-coherence rules. First, we excluded any contact whose nearest above-cutoff neighbor was three or more contacts away, on the grounds that EZ activity manifests as a spatially coherent sEEG pattern rather than a single isolated contact-level abnormality \cite{jaber2024spatial}. Second, we grouped the remaining contacts by electrode, ranked electrodes by their highest-scoring contact, and descended the ranked list, retaining each electrode's contacts until the highest contact score on an electrode dropped by more than a user-defined margin (default 0.2). The remaining contacts constituted the Preliminary EZ predictions and were used to directly evaluate our SFM.

\subsection{Language foundation model}

The language foundation model in EpiiSLM produces the Final EZ predictions with supporting reasoning, informed by multimodal clinical context beyond the signal modality alone. Inputs to the LFM comprise patient-specific information (age, sex, epilepsy type, pathology, and electrode placement); SOZ locations, as determined by neurologists from ictal recordings when available; MRI observations, represented as the absence or location of specific lesion types; and the Preliminary EZ contact predictions and contact-level EZ probability scores produced by the SFM module (Section~\ref{sec:prelim}). Models that incorporate the test patient's SOZ information assume a setting in which seizures were recorded during the monitoring period. We combined these inputs into a single prompt (Supplementary Note \b{11}) and evaluated four pretrained LFMs as candidate backbones: MedGemma \cite{sellergren2025medgemma}, GPT-5.2 \cite{singh2025openai}, GPT-4o \cite{hurst2024gpt}, and MeLLaMA \cite{xie2025medical}. All four were used zero-shot, with no fine-tuning on the MNI or Brno cohorts, and the LFM was kept frozen throughout. We selected MedGemma as the default backbone for its medical-domain pretraining, its zero-shot medical question-answering capability, and its favorable quantitative (Section~\ref{sec:backbones}) and qualitative (Supplementary Note~\b{12}) performance relative to the other candidates. From the LFM's response, we extracted four categories of Final EZ predictions: Core EZ Contacts, Core EZ Regions, Possible Extended EZ Contacts, and Possible Extended EZ Regions. We evaluated Core EZ Contact predictions by PPV ($\frac{\text{TP}}{\text{TP+FP}}$) and Specificity ($\frac{\text{TN}}{\text{TN+FP}}$), and Core EZ Region predictions by region-level accuracy, calculated by comparing identified regions with the surgically resected areas of Engel Class Ia patients (Table~\ref{tab:clinical-characteristics} and  Fig.~\ref{fig:visualization_llm}). Confidence scores reported by EpiiSLM are the contact-level EZ probability scores defined in Section~\ref{sec:prelim}. Full prompts and LFM outputs for all MNI patients appear in Supplementary Note~\b{14}.

We consider the risk of evaluation data contamination in MedGemma minimal. MedGemma's medical fine-tuning data comprise curated medical question--answer, imaging and electronic-health-record datasets \cite{sellergren2025medgemma}, none of which contain our cohort. Patient-level characteristics for this cohort ($n = 30$), including the surgical procedure performed, were published in two previous studies \cite{klimes2019nrem, klimes2022spatio}, and could in principle have entered the general-domain corpus of Gemma 3 (the base model that MedGemma was derived from). To test for memorization directly, we prompted MedGemma to report the surgical procedure for each patient given the patient's published characteristics and an explicit citation to the source; the model reproduced none of the 30 procedures or the table. We performed this analysis for MedGemma only, as it is the backbone of EpiiSLM and produces our headline results; the other candidate LFMs (GPT-5.2, GPT-4o and MeLLaMA) were evaluated solely during backbone selection and are not part of the final model.



\subsection{Implementation details}

We implemented the SFM using a transformer-based encoder. Each sEEG interval was tokenized with a sliding window of 200 samples (hop size 100), producing overlapping tokens. For each token, we computed an FFT over the window (yielding 101 frequency bins) and projected the result through a $101\times256$ fully connected layer to obtain a 256-dimensional token embedding. Token embeddings were processed by a four-layer Linear Attention Transformer \cite{katharopoulos2020transformers} with 8 attention heads, embedding dimension 256, and dropout 0.2 in both attention and post-attention layers.

In Training Phase I, we attached a masked-reconstruction head to the encoder, which consists of a $256\times256$ fully connected layer, GeLU activation, and a second $256\times256$ fully connected layer. This head was used only during Phase I and discarded afterward. Phase I was trained on 8 NVIDIA RTX A5500/A5000/A5000 Ada GPUs with batch size 2048 using the Adam optimizer (learning rate of $10^{-3}$, weight decay of $10^{-5}$) for 100 epochs. In Training Phase II, we initialized both heads from the pretrained encoder and trained them with the following settings. The One-Class head operated on the encoder's output representations directly and was trained with batch size 256, learning rate $10^{-4}$, for 10 epochs. The Binary head comprised an ELU activation followed by a $256\times1$ fully connected layer and was trained with batch size 2048, learning rate $10^{-4}$, for 1 epoch. All Phase II training used the same Adam optimizer. Inference with MedGemma was performed on a single NVIDIA H100 GPU. \s{MOVED to previous section: Contact-level predictions were evaluated by PPV and Specificity, defined as $\frac{\text{TP}}{\text{TP+FP}}$ and $\frac{\text{TN}}{\text{TN+FP}}$, respectively.}

\subsection{Experimental details}

For SFM comparisons (Section~\ref{sec:backbones}), only the SFM encoder architecture was varied. Because SPaRCNet\cite{jing2023development} and STTransformer\cite{song2021transformer} have lower memory efficiency, we reduced the Binary head batch size to 1024 for these models to accommodate GPU memory constraints; all other settings were held constant. For LFM comparisons (Section~\ref{sec:backbones}), only the LFM was varied while the SFM backbone and all other settings remained fixed. For the vigilance-state dependency study (Section~\ref{sec:sleep}), Training Phase II and inference were each performed using data from a single vigilance state (Awake, N1, N2, N3, or REM). Because our cohort contained approximately $9\times$ less N1 than N3 data, we trained the N1 models for proportionally more epochs ($9\times$) to ensure convergence. For the robustness to limited inference data study (Section~\ref{sec:num_cycles}), we simulated reduced N3 availability by restricting inference to consecutive subsets of N3 cycles. We defined one N3 ``cycle'' as 120 consecutive intervals (approximately 20 minutes), consistent with typical N3 cycle durations \cite{altevogt2006sleep}. The three-cycle setting (approximately 60 minutes) was of particular interest because it represents a conservative lower-bound estimate of the N3 sleep typically obtained in a single night. For the clinical prior experiment (Section~\ref{sec:clinical_priors}), full prompt templates for each ablation are provided in Supplementary Note \b{11}.

For the internal distribution-shift experiment (Section~\ref{sec:generalization}), the Inductive setting included only the 15 MNI patients with poor surgical outcomes (Engel Class II–IV) in unsupervised Training Phase I, whereas the Transductive setting included all 30 MNI patients in unsupervised Training Phase I. For the external distribution-shift experiment (Section~\ref{sec:generalization}), Training Phase I included 30 MNI patients in the Inductive setting, 17 Brno patients in the Native setting, and all 47 patients in the Transductive setting. In Training Phase II, the Inductive setting used the 15 MNI Class I patients; the Native setting used the 10 Brno Class I patients under our leave-one-patient-out scheme; and the Transductive setting used all 25 Class I patients, also under leave-one-patient-out. We substituted N2 for N3 data in this experiment because N3 recordings were not available for all Brno Class Ia patients. To equalize the number of forward and backward passes across settings despite differences in dataset size, we trained both heads for $4\times$ the default number of epochs in the Native setting and doubled the batch size in the Transductive setting. We compared performance on the 7 Brno Class Ia patients only. Because the Brno dataset has shorter average inter-contact distances and lacks clinical prior data, we adjusted the user-defined margin (Section~\ref{sec:prelim}) to 0.05.
Unless otherwise specified, all experiments used the following default configuration: BIOT as the SFM backbone, MedGemma as the frozen LFM, N3-only data for Training Phase II and inference, all available multimodal clinical priors, and the Transductive setting.

\subsection{Subgroup and failure-mode analysis}

We assessed performance across patient subgroups and characterized failure cases from the per-patient leave-one-patient-out results (Supplementary Note~\b{3--4}), without additional model runs. Contact-level PPV$_0$ and PPV$_3$ were disaggregated by sex, age (median split), epilepsy type, pathology, and sEEG electrode type across the 10 Engel Class Ia patients, using the patient covariates in Table 1; subgroup summaries are the mean, median and range across patients within each subgroup. Given the cohort size ($n=10$; subgroups $n=1$--$5$), we treated these analyses as exploratory and did not compute per-subgroup confidence intervals or draw conclusions from subgroup hypothesis tests. Failure cases were defined as low contact-level PPV and examined for clustering by the same covariates and by clinical risk. Full results and interpretation appear in Supplementary Note~\b{3--4}.

\subsection{Statistical analysis}

We summarized inter-patient variability using the mean, standard deviation, median, and interquartile range (IQR; Q1--Q3). We computed 95\% confidence intervals (CI) for contact-level PPV$_3$ using the bias-corrected and accelerated (BCa) bootstrap with 10{,}000 resamples, as the distribution of per-patient scores exhibited ceiling effects incompatible with parametric assumptions. We used the Wilcoxon signed-rank test to assess the statistical significance of pairwise model comparisons. To account for stochasticity in model training, we report performance as the mean ± s.d. across three independently seeded runs for our best models EpiiSLM$_{(4)}$ (SFM only) and the full EpiiSLM.

\section*{Ethics declarations}

This study was approved by the Research Ethics Board of McGill University and the Montreal Neurological Institute and Hospital (REB \#2022-8482) and by the Research Ethics Board of St Anne's University Hospital Brno \r{(REB \#39G/2026)}. All participants provided written informed consent prior to sEEG implantation, or consent was obtained under a waiver granted by the relevant ethics committee. The study was conducted in accordance with the Declaration of Helsinki.

\section*{Data availability}

The MNI dataset cannot be publicly shared owing to data size, patient privacy constraints, and institutional ethics board restrictions. Requests for access to de-identified data for non-commercial research purposes can be directed to Professor Birgit Frauscher (birgit.frauscher@mcgill.ca) and are subject to a data use agreement and independent ethics board approval at the requesting institution. The Brno dataset from St Anne's University Hospital Brno similarly cannot be publicly shared owing to data size, patient privacy constraints, and institutional ethics board restrictions; requests may be directed to Professor Milan Brazdil (milan.brazdil@fnusa.cz) and are subject to the same conditions. Summary statistics of patient characteristics for both cohorts are provided in Supplementary Note \b{10}.

\section*{Code availability}

The EpiiSLM codebase and pretrained model weights will be made publicly available upon acceptance. The complete codebase and pretrained model weights are available to editors and reviewers throughout peer review.

\section*{Competing interests}
The authors declare no competing interests.

\section*{Funding}

The AI-based development part of this work was financially supported by the Natural Sciences and Engineering Research Council of Canada (NSERC), Fonds de recherche du Quebec, Canada Foundation for Innovation and the Department of Electrical and Computer Engineering at McGill University. \r{Phenotyping and curation of the MNI dataset was financially supported by a project grant of the Canadian Institutes of Health Research (PJT-175056). Phenotyping and curation of the Brno dataset was financially supported by the Ministry of Health of the Czech Republic (NW25-08-00212).}

\section*{Acknowledgments}

We thank Professor Jean Gotman (Montreal Neurological Institute and Hospital, McGill University) for his contribution to sEEG data collection at MNI and his comments on the manuscript. We thank Professor Laure Peter-Derex (Center for Sleep Medicine and Respiratory Diseases, Lyon University Hospital, Lyon 1 University; Lyon Neuroscience Research Center, Lyon, France) for her contribution to sleep scoring methodology for the MNI dataset and her comments on the manuscript. The authors also wish to acknowledge the partial support of Calcul Quebec and Compute Canada.

\section*{Author contributions}

TKK Ho and T Lai contributed equally to this work and are co-first authors. N Armanfard conceived the study, developed the scientific vision and methodology, and supervised all aspects of the research. N Armanfard is the Principal Investigator and inventor of the proposed approach. TKK Ho, T Lai, and N Armanfard designed the EpiiSLM framework. TKK Ho and T Lai implemented the framework, performed the experiments, and conducted the quantitative analyses under the close supervision and guidance of N Armanfard. B Frauscher, and P Klimes contributed to MNI data collection and curation, and provided valuable feedback on the manuscript through multiple rounds of revision. J Cimbalnik, M Pail, and M Brazdil contributed to Brno data collection and curation. TKK Ho, T Lai, and N Armanfard wrote the manuscript. All authors reviewed the manuscript, provided feedback, and approved the final version.

\section*{Generative AI statement}

 MedGemma \cite{sellergren2025medgemma}, GPT-5.2 \cite{singh2025openai}, GPT-4o \cite{hurst2024gpt}, and MeLLaMA \cite{xie2025medical} were evaluated as candidate LFM backbones during model selection (Section \ref{sec:backbones}). MedGemma was selected as the default LFM in the final EpiiSLM system; GPT-5.2, GPT-4o, and MeLLaMA were used solely as comparative models and are not part of the final system.


\pagebreak

\bibliography{sn-bibliography}

\end{document}